\documentclass[lettersize,journal]{IEEEtran}
\usepackage{amsmath,amsfonts,amssymb}
\usepackage{algorithmic}
\usepackage{algorithm}
\usepackage{array}
\usepackage{textcomp}
\usepackage{stfloats}
\usepackage{url, color, soul}
\usepackage{verbatim}
\usepackage{graphicx}
\usepackage{cite}
\usepackage{hyperref}
\usepackage{amsthm}
\usepackage{multirow}

\usepackage{subcaption}

\newtheorem{lemma}{Lemma}
\newtheorem{theorem}{Theorem}

\newtheorem{corollary}{Corollary}

\newcommand{\R}{\mathbb{R}}

\newcommand{\norm}[1]{\left\lVert#1\right\rVert}

\newcommand{\tr}{\ensuremath{\text{tr}}} 
\newcommand{\var}{\text{Var}}
\newcommand{\RN}{\mathbb{R}^N}
\newcommand{\eye}{\mathbf{I}}

\newcommand{\G}{\mathcal{G}}
\newcommand{\Gm}{\mathcal{G}^m}

\newcommand{\Lapm}{\mathbf{L}^m}
\newcommand{\sigLapm}{{\breve{\mathbf{L}}^m}}
\newcommand{\sigWm}{{\breve{\mathbf{W}}^m}}
\newcommand{\sigGm}{{\breve{\mathcal{G}}^m}}
\newcommand{\Um}{{\mathbf{U}^m}}
\newcommand{\lamnm}{{\lambda_n^m}}

\newcommand{\Wmatm}{\mathbf{W}^m}

\newcommand{\Degmatm}{\boldsymbol{\Gamma}^m}
\newcommand{\Lamm}{\boldsymbol{\Lambda}}
\newcommand{\Lammm}{{\boldsymbol{\Lambda}}^m}

\newcommand{\gj}{{\hat{g}_j}}
\newcommand{\gjpsi}{{\hat{g}_j^{[\psivect]}}}
\newcommand{\gjpsim}{{\hat{g}_j^{[\psim]}}}
\newcommand{\gjpsiz}{{\hat{g}_j^{[\psivect_0]}}}
\newcommand{\gjpsione}{{\hat{g}_j^{[\psione]}}}
\newcommand{\gjpsitwo}{{\hat{g}_j^{[\psitwo]}}}

\newcommand{\Nm}{{N^m}}
\newcommand{\Km}{{K^m}}
\newcommand{\Rm}{{R^m}}
\newcommand{\Rim}{{R_i^m}}
\newcommand{\Nc}{{N}}
\newcommand{\Kc}{{K}}
\newcommand{\Rc}{{R}}
\newcommand{\Ktot}{{K_{T}}}

\newcommand{\Sim}{\mathbf{S}_i^m}
\newcommand{\Sm}{\mathbf{S}^m}
\newcommand{\Simc}{\overline{\mathbf{S}}_i^m}
\newcommand{\Smc}{\overline{\mathbf{S}}^m}
\newcommand{\Qm}{\mathbf{Q}^m}

\newcommand{\yim}{{\mathbf{y}_i^m}}
\newcommand{\yjm}{{\mathbf{y}_j^m}}
\newcommand{\wim}{{\mathbf{w}_i^m}}
\newcommand{\xim}{{\mathbf{x}_i^m}}
\newcommand{\ximest}{{\tilde{\mathbf{x}}_i^m}}
\newcommand{\ximestt}{{\tilde{\mathbf{x}}_i^{m,t}}}
\newcommand{\ximestLast}{{\tilde{\mathbf{x}}_i^{m,T-1}}}
\newcommand{\ym}{{\mathbf{y}^m}}
\newcommand{\xm}{{\mathbf{x}^m}}
\newcommand{\xmest}{{\tilde{\mathbf{x}}^m}}
\newcommand{\wm}{{\mathbf{w}^m}}

\newcommand{\xvect}{{\mathbf{x}}}
\newcommand{\rxvect}{{\underline{\mathbf{x}}}}
\newcommand{\wvect}{{\mathbf{w}}}
\newcommand{\rwvect}{{\underline{\mathbf{w}}}}

\newcommand{\yvect}{{\mathbf{y}}}
\newcommand{\zvect}{{\mathbf{z}}}
\newcommand{\uvect}{{\mathbf{u}}}

\newcommand{\unkyvect}{{\underline{\mathbf{y}}}}

\newcommand{\Ym}{{\mathbf{Y}^m}}
\newcommand{\Xm}{{\mathbf{X}^m}}
\newcommand{\Ymat}{{\mathbf{Y}}}
\newcommand{\Xmat}{{\mathbf{X}}}
\newcommand{\rYmat}{{\underline{\mathbf{Y}}}}
\newcommand{\rXmat}{{\underline{\mathbf{X}}}}

\newcommand{\Amat}{{\mathbf{A}}}
\newcommand{\Bmat}{{\mathbf{B}}}
\newcommand{\Cmat}{{\mathbf{C}}}
\newcommand{\Fmat}{{\mathbf{F}}}
\newcommand{\Nmat}{{\mathbf{N}}}
\newcommand{\Vmat}{{\mathbf{V}}}
\newcommand{\Gmat}{{\mathbf{G}}}
\newcommand{\avect}{{\mathbf{a}}}
\newcommand{\bvect}{{\mathbf{b}}}
\newcommand{\cvect}{{\mathbf{c}}}

\newcommand{\psim}{{\boldsymbol{\psi}^m}}
\newcommand{\psiest}{{\tilde{\boldsymbol{\psi}}}}
\newcommand{\psicom}{{{\boldsymbol{\psi}_C}}}

\newcommand{\delpsi}{{\Delta_{\psi}}}
\newcommand{\delpsisq}{{\Delta^2_{\psi}}}
\newcommand{\psidevm}{{{\boldsymbol{\psi}_{dev}^m}}}
\newcommand{\psione}{{\boldsymbol{\psi}_1}}
\newcommand{\psitwo}{{\boldsymbol{\psi}_2}}
\newcommand{\psivect}{{\boldsymbol{\psi}}}
\newcommand{\Psispace}{{\boldsymbol{\Psi}}}

\newcommand{\thetavect}{{\boldsymbol{\theta}}}
\newcommand{\thetaj}{{\boldsymbol{\theta}_j}}
\newcommand{\rthetavect}{\underline{\thetavect}}

\newcommand{\rpsivect}{\underline{\psivect}}

\newcommand{\xivect}{{\boldsymbol{\xi}}}

\newcommand{\dens}{p}
\newcommand{\meanvect}{\mathbf{m}}
\newcommand{\covmat}{\mathbf{\Sigma}}

\newcommand{\Dict}{{\mathbf{D}}}
\newcommand{\Dm}{{\mathbf{D}^m}}
\newcommand{\Djm}{{\mathbf{D}_j^m}}
\newcommand{\Dmpsi}{{\mathbf{D}^m(\psivect)}}
\newcommand{\Dmpsim}{{\mathbf{D}^m(\psim)}}
\newcommand{\Dmpsiest}{{\mathbf{D}^m(\psiest)}}
\newcommand{\Dmpsicom}{{\mathbf{D}^m(\psicom)}}

\newcommand{\csmooth}{{\mathcal{E}}}
\newcommand{\cparm}{{C_{\Rc}^m}}
\newcommand{\cpar}{{C_{\Rc}}}
\newcommand{\cx}{{C_{x}}}
\newcommand{\cxsq}{{C^2_{x}}}
\newcommand{\cD}{{C_{D}}}
\newcommand{\cDsq}{{C^2_{D}}}
\newcommand{\cprob}{{\tau}}
\newcommand{\cf}{{C_{f}}}
\newcommand{\cpsidiff}{{C_{\psi}}}
\newcommand{\cpsidiffsq}{{C^2_{\psi}}}
\newcommand{\ctesterr}{{\alpha}}
\newcommand{\covnumrate}{{Q}}

\newcommand{\fone}{{h_1}}
\newcommand{\ftwo}{{h_2}}

\newcommand{\errtestm}{{\overline{e}^m}}
\newcommand{\errtestim}{{\overline{e}_i^m}}
\newcommand{\errtrainm}{{{e}^m}}
\newcommand{\errtrainim}{{{e}_i^m}}
\newcommand{\avgerrtrain}{{Z}}

\newcommand{\dist}{{{d}}}
\newcommand{\covcents}{{\mathcal{C}}}
\newcommand{\covnumeps}{{\mathcal{N}_\epsilon}}
\newcommand{\covnumcprobeps}{{\mathcal{N}_{\cprob \varepsilon}}}

\begin{document}

\title{Graph Signal Inference by Learning Narrowband Spectral Kernels}

\author{Osman Furkan Kar, G\"ulce Turhan and Elif Vural
\thanks{The authors are with the Department of Electrical-Electronics Engineering, METU, Ankara. This work was supported by the Scientific and Technological Research Council of Turkey (T\"UB\.ITAK) under grant 120E246.}}




\maketitle

\begin{abstract}

While a common assumption in graph signal analysis is the smoothness of the signals or the band-limitedness of their spectrum, in many instances the spectrum of real graph data may be concentrated at multiple regions of the spectrum, possibly including mid-to-high-frequency components. In this work, we propose a novel graph signal model where the signal spectrum is represented through the combination of narrowband kernels in the graph frequency domain. We then present an algorithm that jointly learns the model by optimizing the kernel parameters and the signal representation coefficients from a collection of graph signals. Our problem formulation has the flexibility of permitting the incorporation of signals possibly acquired on different graphs into the learning algorithm. We then theoretically study the signal reconstruction performance of the proposed method, by also elaborating on when joint learning on multiple graphs is preferable to learning an individual model on each graph. Experimental results on several graph data sets shows that the proposed method offers quite satisfactory signal interpolation accuracy in comparison with a variety of reference approaches in the literature. 

\end{abstract}


\section{Introduction}
\label{sec_intro}

\IEEEPARstart{L}{earning} efficient graph signal models is a key problem of interest for the analysis and inference of network data. Graph signals in real-world applications often exhibit dominant frequency components across different parts of the spectrum. For example, in a social network where user behavior is considered as a graph signal, various factors inherited from a broader community of users result in slowly-varying low-frequency components, whereas other factors affecting smaller social groups such as friend circles may manifest as abrupt changes in user behavior on the graph and produce significant mid- or high-frequency components in signal spectra.  Similarly, in a meteorological sensor network, global seasonal characteristics such as cold weather in winter are associated with low-frequency components, while sharp local temperature variations in micro-climate zones give rise to high-frequency components in the spectrum of measured signals.

Motivated by these observations, in this paper, we study the problem of learning graph signal models where signals are represented in terms of a combination of narrowband spectral kernels, with the purpose of accurately capturing their frequency characteristics in different spectral regions. We demonstrate the usage of our algorithm in signal interpolation applications, i.e., for estimating the missing entries of partially observed graph signals, which is of interest in  a variety of applications where data is not available over all graph nodes due to issues such as sensor failure, connection loss, or unavailable user information. 

\begin{figure}
     \centering
     \begin{subfigure}[b]{0.2\textwidth}
         \centering
         \includegraphics[width=\textwidth]{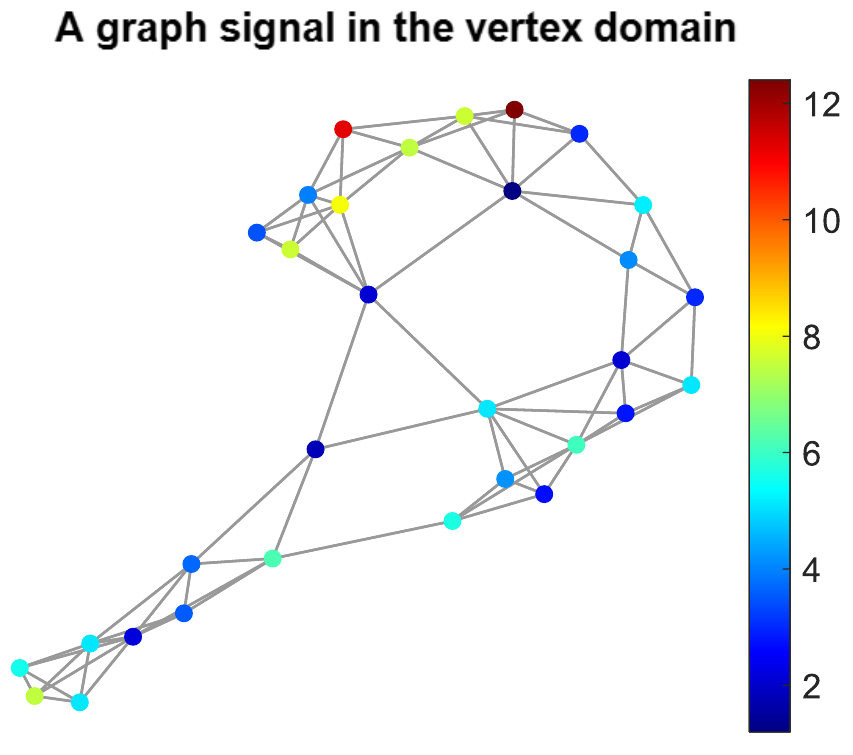}
         \caption{}
     \end{subfigure}
     \hfill
     \begin{subfigure}[b]{0.2\textwidth}
         \centering
         \includegraphics[width=\textwidth]{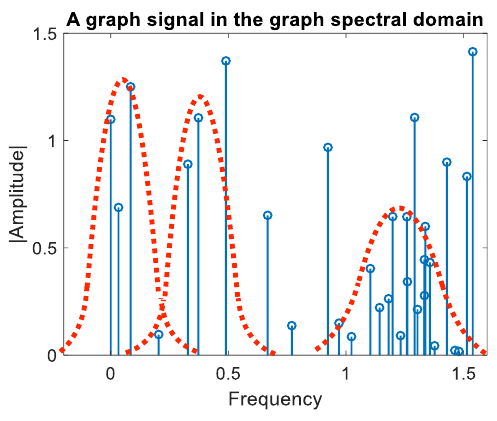}
         \caption{}
         \label{fig_molene_signal_spectrum}
     \end{subfigure}
     \hfill
	\caption{(a) A real graph signal consisting of wind speed measurements from the Mol\`{e}ne data set \cite{Girault-Stationarity} (b) The signal spectrum  contains middle and high frequencies in addition to low frequencies (blue). Our approach is based on fitting narrowband kernels to high-energy spectral components (red).}
 \label{fig_molene_signal}
\end{figure}

With the growing literature on graph signal processing, a wide scope of solutions now exists for the inference and modeling of graph signals. Graph neural network (GNN) models offer state-of-the-art solutions in various tasks including node classification, link prediction, graph embedding and graph signal reconstruction \cite{ZonghanPCLZY21}. Attention mechanisms have been integrated into GNN architectures in the recent years, leading to the popular graph attention network (GAT) models \cite{VelickovicCCRLB18, Brody0Y22}. However, one limitation of graph neural networks is that they typically need larger data sets compared to traditional inference methods due to the relatively large number of model parameters they involve. Also, well-known effects such as oversmoothing \cite{ZhaoA20} and vulnerability to imbalance and adversarial attacks \cite{ZhangZ2020} may play a negative role on their performance. In contrast to the complexity of GNN architectures, classical reconstruction approaches are also still widely favored due to their appealing complexity-performance tradeoff in signal interpolation applications, which include graph-based regularization methods \cite{JungHMJHE19}, non-smooth interpolation techniques \cite{MAZARGUIL2022108480}, and reconstruction approaches based on band-limited models \cite{8918094, LorenzoBIBL18, YangYYH21}. These methods often rely on rather strict assumptions about signal structures, such as signal self-similarity, band-limitedness, or low-pass behavior due to smoothness priors on the graph. On the other hand, actual graph signals encountered in real life do not necessarily conform to such low-frequency or band-limited models. This is demonstrated in Figure \ref{fig_molene_signal}, where the spectrum of a meteorological graph signal is plotted. One can clearly observe that the signal spectrum concentrates not only at low frequencies, but also in the mid-to-high frequency range. 

In order to address these considerations, in this work we adopt a graph signal model where signals are represented over a set of graph signal prototypes generated from a set of narrowband kernels. We formulate an optimization problem where the kernel parameters are learnt so as to model different spectral components of graph signals as illustrated in Figure  \ref{fig_molene_signal_spectrum}. In addition to the regularization of the kernel parameters and the rate of change of graph signals, our objective function also includes a third regularization term imposing that similar graph signals have similar representations. Especially in settings where graph signals are only partially observed, this third term proves useful for fusing the information in different signals in order to arrive at an accurate model. The resulting optimization problem is solved with an alternating optimization approach and the initially missing observations of the signals are computed based on the learnt model. The proposed method involves the learning of relatively few model parameters in the spectral domain, i.e., only the center and scale parameters of kernels. This lightweight structure of our model provides robustness to the limited availability of training data, in comparison with more complex schemes that require the learning of a large number of model parameters.  


Since we learn graph models in the frequency domain, the learnt spectral model can be extended to any given graph. This allows the incorporation of graph signal data coming from multiple independently constructed graphs in the learning, which is especially interesting when the graphs have similar topologies and host similar types of signals, e.g.,~as in an application where similar traffic statistics are acquired on the transportation networks of cities of comparable size. The idea of learning from multiple graphs has already been explored in several previous studies \cite{ThanouF15, SindhwaniR08}. An important question is under which conditions the incorporation of multiple graphs into the algorithm impacts the learning performance positively. With the aim of answering this question, we then present a theoretical analysis of the performance of our algorithm. We first propose performance bounds on the estimation error of our method in terms of the number $M$ of graphs and data size $\Kc$, which suggest that the reconstruction error probability decreases at rate $O(1/(M \Kc))$ as the number of graphs and data size increases. We then compare the regimes of joint learning on multiple graphs and individual learning on single graphs. Our theoretical results show that joint learning turns out to be advantageous when the spectral discrepancy $\delpsisq $ between the signals on multiple graphs is sufficiently low in relation to the data size $\Kc$, pointing to an inverse quadratic relation $\Kc < O(1/ {\delpsisq})$ between their critical values.

In Section \ref{sec_related_work}, we overview the related work. In Section \ref{sec:ouralgorithm}, we introduce our graph signal model and in Section \ref{sec:prop_method}, we propose a method for learning the model. In Section \ref{sec_theo_anlys}, we present a theoretical analysis of our algorithm and study the trade-off between the multi-graph spectral discrepancy and data size. In Section \ref{sec_exp_results}, we evaluate the performance of our method with experiments and in Section \ref{sec_conclusion}, we conclude.

\section{Related Work}
\label{sec_related_work}

The reconstruction of partially observed graph signals is a well-studied problem in the literature and many different solution approaches exist. Classical approaches include Tikhonov regularization \cite{shuman2013emerging}, piecewise constant or piecewise planar signal models \cite{JungHMJHE19, ChenCZ24}, kernel-based methods \cite{JianTE24}, non-smooth graph signal interpolation \cite{MAZARGUIL2022108480, ioannidis2019semi}, iterative graph signal reconstruction methods \cite{narang2013localized, wang2015local, christensen2003frames, 8918094}, and techniques based on the bandlimitedness assumption \cite{8918094,  LorenzoBIBL18, YangYYH21, WangCYLF23}. Each of these approaches employs a particular strategy or prior in order to effectively utilize the graph signal information. Traditional Tikhonov regularization methods, optimal iterative reconstruction (O-PGIR) \cite{8918094}, and optimal sampling  \cite{LorenzoBIBL18, YangYYH21} strategies rely on low-pass or band-limited signal models, which may fall short of capturing band-pass or high-pass variations in signal spectra as demonstrated in Fig.~\ref{fig_molene_signal},  especially when signals display isolated and localized behaviors in specific graph regions. TV-regularization ideas \cite{JungHMJHE19} and affine signal models as in \cite{MAZARGUIL2022108480} may better handle non-smooth signal variations; however, they rely on other constraints such as piecewise constancy or low-rank structure of graph signals, which may not always be met in practice.



Recent trends in the analysis of graph data focus considerably on graph neural network (GNN) models \cite{defferrard2016convolutional, kipf2017semi, HamiltonYL17}, with many variants building on spectral approaches \cite{levie2019cayleynets}, diffusion or message passing models \cite{atwood2016diffusion, Brockschmidt20, ZouCZLZWL24}, and graph attention models \cite{VelickovicCCRLB18, Brody0Y22}.  However, GNNs are subject to two significant limitations: a lack of interpretability and reliance on relatively large training datasets. Additionally, deeper networks do not consistently translate to improved performance within graph settings \cite{ZhaoA20}, in contrast to the notable success of convolutional networks for signals on regular grids. Given the aforementioned limitations of GNNs, deep algorithm unrolling methods introduce a hybrid approach between traditional graph regularization schemes and GNN-based restoration methods, incorporating learnable parameters into iterative algorithms \cite{chen2021graph, nagahama2022graph}.

Our method relies on learning narrowband graph signal prototypes for the reconstruction of graph signals, which bears resemblance to a graph dictionary learning problem. Several previous works have studied graph signal representations over predetermined or learnt graph dictionaries. The  spectral graph wavelet dictionaries (SGWT) proposed in \cite{hammond2011wavelets} extends the wavelet theory to graph domains in view of their spectral characterization. Other efforts include the learning of parametric graph dictionaries on single graphs \cite{thanou2014learning} and multiple graphs \cite{thanou2015multi}, as well as multi-scale graph dictionaries based on Haar wavelets \cite{8642839}. Although our method can be interpreted as a particular type of graph dictionary learning algorithm, it has the following essential differences from the above methods. First, the aforementioned methods are generic dictionary learning algorithms that require fully observed graph signals to train; hence, are not particularly suited to the graph signal reconstruction problem and lack the capability of learning models with only partially observed graph signals. Similarly, they neither  employ any priors on the spectral characteristics of the data, nor present a theoretical understanding of their reconstruction performance. In contrast, our solution actively employs the prior that the signal energy is concentrated in certain bands of the spectrum, thanks to which it involves relatively few model parameters to learn (limited to the center and scale parameters of the kernels fitting to the dominant spectral components), hence, it is particularly tailored to scenarios with severe lack of training data. This feature of our method also makes it favorable against sophisticated but more complex methods such as GNN-based solutions requiring the learning of a large set of model parameters.  Furthermore, we provide an extensive theoretical analysis of the signal reconstruction performance of our method, with a careful justification of when multi-graph learning is advantageous over individual learning.


Preliminary versions of our study have been presented in \cite{TurhanV21, KarTV22}. The current paper builds on these studies by significantly extending the experimental results and including a detailed theoretical analysis.

\section{Proposed Graph Signal Model}
\label{sec:ouralgorithm}

We consider a setting with independently constructed multiple graphs $\mathcal{G}^1,\mathcal{G}^2, \dots, \mathcal{G}^M$, where each graph $\mathcal{G}^m = (\mathcal{V}^m ,\Wmatm)$ is undirected and weighted, with vertices $\mathcal{V}^m$, edge weight matrix $ \Wmatm \in \R^{\Nm \times \Nm}$,  for $ m=1,2, \dots, M $.  We denote the number of nodes (vertices) on graph $\mathcal{G}^m$ as $ N^m $. The normalized graph Laplacian of each graph is defined as
\begin{equation*}
	\Lapm=(\Degmatm)^{-1/2} (\Degmatm - \Wmatm) (\Degmatm)^{-1/2} 
\end{equation*}
where $\Degmatm \in \R^{\Nm \times \Nm}$ is the diagonal degree matrix with diagonal entries $\Degmatm_{ii}= \sum_ j \Wmatm_{ij}$. The eigenvector matrix $\Um$ of the graph Laplacian $\Lapm $ provides a graph Fourier basis for  graph signals, where $ \Lapm =\Um \Lammm(\Um)^T $ is the eigenvalue decomposition of $\Lapm $.

 
\subsection{Graph Dictionary Model}

We propose to represent graph signals in terms of narrowband graph signal prototypes, which form a graph dictionary that is designed capture the dominant spectral components of the signals. In the previous works \cite{thanou2014learning}, \cite{thanou2015multi}, the graph dictionary learning problem has been formulated in the spectral domain through the use of a set of spectral graph kernels $ \hat{g}_j(\lambda) : \mathbb{R} \rightarrow  \mathbb{R}$,  for $ j=1,2, \dots, J $, each of which is a function of the graph frequency variable $\lambda$ in the spectral domain.  The application of the kernel $ \hat{g}_j(\lambda)  $ to the graph Laplacian $\Lapm $ generates a subdictionary \cite{thanou2014learning}, \cite{thanou2015multi}
\[ \Djm =\Um \hat{g}_j(\Lammm)(\Um)^T \in \mathbb{R}^{N^m\times N^m}\]
where $\hat{g}_j(\Lammm)$ is a diagonal matrix obtained by applying the kernel  $ \hat{g}_j(\cdot)  $ to eigenvalues $\{ \lambda_n^m \}$ of the graph Laplacian $\Lapm$ found in matrix $\Lammm$. Here, each $n$-th column of the subdictionary $\Djm$ is an atom that gives the graph signal obtained by localizing (centralizing) the graph kernel $ \hat{g}_j(\lambda)  $ at node $ n $. A structured dictionary is then obtained by concatenating $ J $ subdictionaries as
\begin{equation*}
	 \Dm=\begin{bmatrix} 
		\Dict_1^m & \Dict_2^m & \cdots & \Dict_J^m
	\end{bmatrix}  \in  \mathbb{R}^{ N^m\times J N^m}.
\end{equation*}

Our aim in this work is to learn graph dictionaries that can successfully capture different spectral components of the observed graph signals at hand. Given the lack of data in our particular setting with possibly partially observed graph signals, in order to keep the number of dictionary model parameters small while efficiently fitting the atoms to the data spectrum, we propose to choose the spectral kernels $ \hat{g}_j(\lambda) $ as Gaussian kernels, whose center frequency and frequency spread can be controlled in a way to represent the signal set accurately. We thus express each $ \hat{g}_j(\lambda) $ in a parametric form with the associated parameter vector $ \thetaj=(\mu_j,s_j )$ as 
\begin{equation*}
\hat{g}_j(\lambda)=\exp \left(- \frac{\norm{\lambda-\mu_j}^2}{s_j^2} \right)
\end{equation*}
where $ \mu_j $ and  $ s_j $ represent the mean and the scale parameters of the kernel. 
We propose to learn a total of $J$ Gaussian kernels $\hat{g}_1(\lambda), \cdots, \hat{g}_J(\lambda)$ that are common for all graphs $\mathcal{G}^m$. In this way, each Gaussian kernel  $ \hat{g}_j(\lambda) $ is designed to generate graph signal prototypes that fit to a different spectral component of the partially known graph signals at hand. Our method is capable of fusing the information of incomplete observations of the signals from all graph domains when learning the kernel parameters, which is expected to improve the performance under certain conditions.


%
%
%

\subsection{Graph Signal Model}
\label{ssec_graph_signal_model}

A graph signal $\ym \in \mathbb{R}^{N^m}$ on graph $\mathcal{G}^m$  is a real vector, where each $n$-th entry $\ym(n)$ of the signal represents its value at graph node $n$. We model each $i$-th graph signal $ \yim  \in \mathbb{R}^{N^m}$ on graph  $\mathcal{G}^m$  as
\begin{equation}
\begin{split}
\label{eq_signal_model}
	\yim&=\begin{bmatrix} 
		\Dict_1^m & \Dict_2^m& \cdots & \Dict_J^m
	\end{bmatrix} \xim+ \wim\\
	&=\Dmpsi \, \xim + \wim.
\end{split}
\end{equation}
Here $ \psivect=[\mu_1 \ \cdots \ \mu_J \ s_1 \ \cdots \  s_J ]$ denotes the overall kernel parameter vector consisting of the Gaussian kernel parameters of different spectral components, such that $ \Dmpsi $ represents the graph dictionary on graph $\mathcal{G}^m$ generated by $\psivect $. The vector $ \xim  $ consists of the coefficients for $\yim$, and $ \wim $ denotes a noise signal.  

In the signal model \eqref{eq_signal_model}, the kernel parameter vectors $ \thetaj $ generating the subdictionaries $\{ \Djm \}$ are assumed to be independently sampled from a normal distribution $ \mathcal{N}(\meanvect_{\rthetavect},\covmat_{\rthetavect})\ $ with density $ \dens_{\rthetavect}(\thetavect) $, where $ \meanvect_{\rthetavect} $ is the mean vector and $\covmat_{\rthetavect}$ is a diagonal covariance matrix. Next, in order to encourage the sparsity of the coefficients, the coefficients in vectors $ \xim $ are modeled to be independently sampled from a Laplace distribution with parameter $\delta$, with joint density $ \dens_{\rxvect}(\xvect)$. Finally, the noise signals $ \wim $ are assumed to be independently sampled from a zero-mean normal distribution $ \mathcal{N}(0,\sigma^{2} \, \eye_\Nm )\ $ with density $ \dens_{\rwvect}(\wvect) $, where  $\eye_{N} \in \R^{N\times N}$ denotes the identity matrix.

Let us define the graph signal matrix $\Ym = [\yvect_1^m \ \dots \ \yvect_{K^m}^m] \in \mathbb{R}^{N^m \times K^m}$ consisting of $K^m$ graph signals observed over the graph $\mathcal{G}^m$, and the  corresponding coefficient matrix $\Xm = [\xvect_1^m \ \dots \ \xvect_{K^m}^m] \in \mathbb{R}^{J N^m \times K^m}$.  Assuming the independence of $\Xm $,  $ \psivect$, and the noise signals $\{ \wim \}$, the joint density of  the overall signal matrix $ \Ymat=[\Ymat^1 \ \cdots \ \Ymat^M] $, the parameter vector $\psivect$, and the overall coefficient matrix  $ \Xmat= [\Xmat^1 \ \cdots \ \Xmat^M] $  is given by 
\begin{eqnarray}
\label{eq:joint_distYXpsi}
	\dens_{\rYmat,\rpsivect,\rXmat}(\Ymat,\psivect, \Xmat)&=& \dens_{\rpsivect}(\psivect) 
	\ 
	\dens_{\rXmat |\underline{\psivect}}( \Xmat | \psivect) 
	\
	\dens_{\rYmat | \rXmat ,\rpsivect}( \Ymat | \Xmat, \psivect) \nonumber \\
	&=& \dens_{\rpsivect}(\psivect) 
	\
	\dens_{\rXmat}(\Xmat) 
	\
	\dens_{\rYmat | \rXmat,\rpsivect} (\Ymat | \Xmat, \psivect)
\end{eqnarray}
where
\begin{eqnarray*}
			&\dens_{\rpsivect}(\psivect)=\prod_{j=1}^{J}\dens_{\rthetavect}(\thetaj)\nonumber \\
			&=\prod_{j=1}^{J}\frac{1}{2\pi \sqrt{| \covmat_{\rthetavect} |}} 
			\exp\left(-\frac{1}{2}(\thetaj- \meanvect_{\rthetavect})^T {\covmat_{\rthetavect}}^{-1}(\thetaj- \meanvect_{\rthetavect})\right)
\end{eqnarray*}
and
\begin{eqnarray*}
	\dens_{\rXmat}(\Xmat) &=&\prod_{m=1}^{M}\prod_{i=1}^{K^m} 
	\dens_{\rxvect}(\xim) \nonumber \\ 
	&=&\prod_{m=1}^{M}\prod_{i=1}^{K^m}\prod_{l=1}^{JN^m}\frac{1}{2\delta} 
	\exp\left(-\frac{| \xim (l)|}{\delta}\right) \nonumber\\
	&=&\prod_{m=1}^{M}\prod_{i=1}^{K^m}\frac{1}{(2\delta)^{JN^m}} 
	\exp\left(-\frac{{\| \xim \|}_1}{\delta}\right)\nonumber \\
	&=&\frac{1}{(2\delta)^{J(\sum_{m=1}^{M}N^m K^m)}} \exp\left(-\frac{{\| \Xmat \|}_1}{\delta}\right) \nonumber .\\
\end{eqnarray*}
Here $\| \cdot \| $ denotes the $\ell_1$-norm of a vector or the vectorized form of a matrix. The conditional density of the signal matrix $\Ymat$ is obtained similarly as
\begin{equation*}
\begin{split}
	&\dens_{\rYmat | \rXmat,\rpsivect} (\Ymat | \Xmat, \psivect)
	=\prod_{m=1}^{M}\prod_{i=1}^{\Km}
	\dens_{\rwvect}(\yim- \Dmpsi \xim)\nonumber \\
	&=\prod_{m=1}^{M}\prod_{i=1}^{\Km}\frac{1}{\sqrt{(2\pi)^{N^m}\sigma^{2N^m}}}
	\exp \left(-\frac{1}{2\sigma^2}\| \yim-\Dmpsi \xim\|^2\right)  \\
	&=\frac{1}{(2\pi\sigma^2)^{\sum_{m} \frac{\Nm \Km}{2}}} 
	\exp\left(-\frac{1}{2\sigma^2}\sum_{m=1}^{M} \| \Ym-\Dmpsi \Xm\|_F^2\right)\nonumber 
\end{split}
\end{equation*}
where $\| \cdot \|_F$ denotes the Frobenius norm of a matrix.

\section{Proposed Method for Learning Graph Signal Representations}
\label{sec:prop_method}

\subsection{Problem Formulation}


We assume a flexible setting where each signal $ \yim \in \mathbb{R}^{N^m} $ on graph $\mathcal{G}^m$ may be observed partially;  i.e., only at $R_i^{m}$ nodes out of $N^m$ nodes. Let us denote by $n_{i,k}^{m}$ the indices corresponding to the observed (available) entries of each signal $ \yim$, for $k=1,2,\dots,R_i^{m}$. In order to extract the observed entries of each signal  $\yim$, let us also define a selection matrix $ \Sim \in \mathbb{R}^{R_i^m \times N^m}$ whose entries indexed with $(k, n^m_{i,k})$ are $1$, and the remaining entries are $0$. 
In this setting, the maximization of the log likelihood of the distribution in \eqref{eq:joint_distYXpsi} is then equivalent to the optimization problem 
\begin{equation}	
	\begin{split}
		&\min \limits_{\{\Xm\},\psivect}  \ \ \eta_{\theta}\sum_{j=1}^{J}
		(\thetaj- \meanvect_{\rthetavect})^T {\covmat_{\rthetavect}}^{-1}(\thetaj- \meanvect_{\rthetavect})
		+\eta_{x}\sum_{m=1}^{M}\norm{\Xm }_1 \\
		&+\eta_{w}\sum_{m=1}^{M}\sum_{i=1}^{K^m}\norm{\Sim \yim-\Sim \Dmpsi \xim}^2
	\end{split}
\label{eq:init_obj}
\end{equation}
where the positive weight parameters $ \eta_{\theta} $, $ \eta_{x} $ and $ \eta_{w} $ come from the constants of the distributions.

Assuming independent distributions $\mu_j \sim \mathcal{N}(0,\sigma_{\mu}^2) $, ${s_j} \sim \mathcal{N}(s_0,\sigma_{s}^2)$ for the mean and the scale parameters of the Gaussian kernels,  
%
the MAP estimation problem in \eqref{eq:init_obj} can be rewritten as

\begin{equation}
	\begin{split}
		&\min \limits_{\{\Xm\},\psivect} \ 
		   \sum_{j=1}^{J}(\mu_j)^2+\eta_{s}\sum_{j=1}^{J}(s_j- s_0)^2+\eta_{x}\sum_{m=1}^{M}\norm{\Xm}_1\\
		&+\eta_{w}\sum_{m=1}^{M}\sum_{i=1}^{K^m}\norm{\Sim \yim-\Sim \Dmpsi  \xim}^2 
	\end{split}
	\label{eq_obj_final_map}
\end{equation}
where $\eta_s$ is a positive weight parameter. The first and the second terms in the above objective function can be regarded as regularization terms where the first term prevents the center frequencies $\mu_j$ of the Gaussian kernels from shifting to too high frequencies, and the second term draws the kernel bandwidth parameters $s_j$ towards a predetermined nominal value $s_0$. The third term encourages the sparsity of the representation coefficients, while the fourth term enforces the learnt graph signal model to be coherent with the available observations of the signals $\yim$.  While the formulation in \eqref{eq_obj_final_map} is insightful as it originates from a MAP estimation problem, we propose to incorporate two additional regularization terms in the objective function in an effort to improve the signal estimation accuracy as follows:
\begin{equation}	
	\begin{split}
		&\min \limits_{\{\Xm\},\psivect} \  \ \sum_{j=1}^{J}(\mu_j)^2+\eta_{s}\sum_{j=1}^{J}(s_j- s_0)^2+\eta_{x}\sum_{m=1}^{M}\norm{\Xm }_1 \\
		&+\eta_{w}\sum_{m=1}^{M}\sum_{i=1}^{\Km}\norm{\Sim \yim- \Sim \Dmpsi \xim}^2\\
		&+\eta_{y}\sum_{m=1}^{M} \tr(( \Xm)^{T}( \Dmpsi )^{T} \Lapm  \Dmpsi \Xm)\\
		&+\eta_{c}\sum_{m=1}^{M} \tr(( \Xm) \sigLapm ( \Xm)^{T})
	\end{split}
\label{eq_obj_final_reg_map}
\end{equation}
In this final form of the objective function, the regularization term weighed by the parameter $\eta_y>0$ encourages the signals  $ \Dmpsi \Xm$ reconstructed from the learnt model to vary smoothly on the graph $\mathcal{G}^{m}$. Meanwhile, the last regularization term weighed by $\eta_c>0$ aims to ensure that similar signals to be reconstructed on each  $\mathcal{G}^{m}$  use a similar set of graph signal prototypes in their representations: The similarity between a graph signal pair $(\yim, \yjm)$ is identified through a signal affinity matrix $ \sigWm \in \mathbb{R}^{K^m \times K^m}$ with entries given by
\begin{equation}	
	\begin{split}
		&\sigWm_{ij} = \exp( - \norm{\Qm_{i,j}  \yim - \Qm_{i,j} \yjm}^2 / \gamma^2).
	\end{split}
\label{eq:init_obj3}
\end{equation}
Here,  $\Qm_{i,j}$ are selection matrices that consist of 0's and 1's, which extract the node indices where both of the signals in the pair $(\yim, \yjm)$ are observed. 
The parameter $\gamma$ is a suitably chosen scale parameter. In fact, it is possible to regard the matrix $ \sigWm$ in \eqref{eq:init_obj3} as the weight matrix of a ``signal graph'' $\sigGm$ with $K^m$ nodes, whose $i$-th node stands for the signal $\yim$, and whose edge weights model the affinities between signal pairs $(\yim, \yjm)$. The matrix  $\sigLapm \in \mathbb{R}^{K^m \times K^m}$ appearing in the last term in the objective \eqref{eq_obj_final_reg_map} then represents the Laplacian of the graph  $\sigGm$ formed in this way.  Hence the last term in \eqref{eq_obj_final_reg_map} enforces each row of the coefficient matrix $\Xm$ to have a slow variation on the signal graph  $\sigGm$, so that  graph signals with high affinity have similar representations in the learnt model. As different graph signals will often have different missing entries in practice, the proposed regularization approach is useful for efficiently fusing the information available in all individual signals in order to arrive at a globally coherent graph signal model that accurately captures the common characteristics of the signal collection at hand.

\subsection{Solution of the Optimization Problem}
\label{ssec_opt_prob_soln}

The objective function of the optimization problem \eqref{eq_obj_final_reg_map} is not jointly convex in the coefficient matrices $\{ \Xm\}$ and the parameter vector $\psivect$; nevertheless, it is convex when considered as a function of only  $\{ \Xm\}$. We minimize the objective function with an iterative alternating optimization approach. In each iteration, we first optimize the coefficients $\{ \Xm \}$ by fixing $\psivect$, and then optimize $\psivect$ by fixing $\{ \Xm \}$. We discuss below the details of the proposed optimization procedure.

\subsubsection{Optimization of the Coefficients $ \Xm $}
\label{ssec:OptXm}

Fixing the parameter vector $\psivect$, the problem \eqref{eq_obj_final_reg_map} becomes  
\begin{equation}	
	\begin{split}
		&\min\limits_{\{ \Xm \}}\eta_{x}\sum_{m=1}^{M}\norm{\Xm }_1 
		+\eta_{w}\sum_{m=1}^{M}\sum_{i=1}^{K^m}\norm{ \Sim \yim- \Sim \Dmpsi  \xim}^2\\
		&+\eta_{y}\sum_{m=1}^{M}\tr(( \Xm)^{T}(\Dmpsi)^{T} \Lapm \Dmpsi \Xm)\\
		&+\eta_{c}\sum_{m=1}^{M}\tr(( \Xm) \sigLapm ( \Xm)^{T})
				\label{eq:init_obj4} 
	\end{split}
\end{equation}
which can be rewritten as
\begin{equation}	
	\begin{split}
		&\min \limits_{\{ \xim \}} \ \eta_{x}\sum_{m=1}^{M}\sum_{i=1}^{ \Km}\norm{ \xim }_1 \\
		&+\eta_{w}\sum_{m=1}^{M}\sum_{i=1}^{\Km}\norm{ \Sim \yim- \Sim \Dmpsi \xim}^2\\
		&+\eta_{y}\sum_{m=1}^{M}\sum_{i=1}^{\Km}(\xim)^{T}(\Dmpsi)^{T} \Lapm \Dmpsi \xim\\
		\label{eq:init_obj5}
		&+\eta_{c}\sum_{m=1}^{M}\tr(( \Xm) \sigLapm ( \Xm)^{T}).
	\end{split}
\end{equation}
Although the problem \eqref{eq:init_obj5} is jointly convex in the coefficient matrices $\{ \Xm \}$, it may be too complex to globally optimize the matrices $\{ \Xm \}$ especially when the number of graph signals is large. We thus propose an approximative solution to the problem \eqref{eq:init_obj5} where we minimize the objective iteratively, such that in each iteration we optimize only one  coefficient vector $\xim$ and fix the others. Regarding the objective function as a function of only $\xim $, we minimize it with the ADMM algorithm as follows \cite{BoydPCPE11}: Defining the auxiliary vectors $\zvect \in \R^{J N^m \times 1}$ and $\uvect \in \R^{J N^m \times 1}$, and denoting the optimization variable as $\xvect \triangleq \xim \in \R^{J N^m \times 1}$ for simplicity, the augmented Lagrangian of the problem is given by
\begin{equation}	
	\begin{split}
		&L_\rho( \xvect, \zvect, \uvect)=\eta_{w}\norm{ \Sim \yim- \Sim \Dmpsi \xvect}^2\\
	&+\eta_{y} \, \xvect^{T}(\Dmpsi)^{T} \Lapm \Dmpsi \xvect
	+\eta_{c} \, \tr(( \Xm) \sigLapm ( \Xm)^{T})\\
	&+\eta_{x}\norm{\zvect}_1+\rho/2\norm{\xvect- \zvect+\uvect}^2.
	\end{split}
\label{eq:init_obj6}
\end{equation}
where $\rho>0$ is a penalty parameter. Grouping the terms depending on only $\xvect$ and only $\zvect$ in the functions
\begin{equation}	
	\begin{split}
		&g(\xvect)=\eta_{w}\norm{ \Sim \yim- \Sim \Dmpsi \xvect}^2 \\
		&+\eta_{y} \, \xvect^{T}(\Dmpsi )^{T} \Lapm \Dmpsi \xvect 
		+\eta_{c} \, \tr(( \Xm) \sigLapm ( \Xm)^{T})\\
&\text{ and }\quad	h(\zvect)=\eta_{x}\norm{\zvect}_1,
	\end{split}
\label{eq:init_obj7}
\end{equation}
the ADMM updates are obtained as \cite{BoydPCPE11}
\begin{eqnarray}	
		&\xvect^{k+1}&=\arg \min \limits_{\xvect} g(\xvect)+\rho/2\norm{\xvect -\zvect^k+ \uvect^k}^2 \label{eq_obj8_xupd}\\
		&\zvect^{k+1}&=\arg \min\limits_{\zvect} h(\zvect)+\rho/2\norm{\xvect^{k+1}- \zvect+ \uvect^k}^2
\label{eq_obj8_zupd}\\
		&\uvect^{k+1}&=\xvect^{k+1}-\zvect^{k+1}+ \uvect^k \label{eq_obj8_uupd}. 
\end{eqnarray}
Here as $g(\xvect)$ is a quadratic function of $\xvect$, the optimal $\xvect^{k+1}$ vector can be found analytically by setting the gradient of the objective \eqref{eq_obj8_xupd} to 0 as follows: The gradients of the terms except for the term  $\eta_{c} \, \tr((\Xm) \sigLapm ( \Xm)^{T})$ in \eqref{eq_obj8_xupd} involve expressions in terms of the matrices 
\begin{equation}	
	\begin{split}
		& \Amat=\eta_{w} \, (\Dmpsi)^{T} ( \Sim )^{T} \Sim \Dmpsi \\
		&+\eta_{y} \, (\Dmpsi)^{T} \Lapm \Dmpsi
		+\rho/2 \, \eye_{JN^m}\\
		\text{ and } 
		&\bvect=\rho( \uvect^k- \zvect^k)-2\eta_{w} \, (\Dmpsi)^{T}(\Sim )^{T} \Sim \yim. \\
	\end{split}
\label{eq:init_obj9}
\end{equation}
In addition, defining the matrix  $\sigLapm_d \in \mathbb{R}^{K^m \times K^m}$ which consists of only the diagonal elements of  $\sigLapm$, one can decompose the ``signal graph'' Laplacian as $\sigLapm  = \sigLapm_d + \sigLapm_o$, where the matrix $ \sigLapm_o \in \mathbb{R}^{K^m \times K^m}$ consists of the off-diagonal entries of $\sigLapm $. In this case we have the equality
\begin{equation}	
	\begin{split}
		& \eta_{c} \tr( \Xm \sigLapm ( \Xm)^{T}) =
		 \eta_{c} \tr( \Xm (\sigLapm_d + \sigLapm_o )( \Xm)^{T}) \\
		&= \eta_{c} \bigg(\tr(\Xmat_i^{m} \sigLapm_d (\Xmat_i^{m})^{T})  \\
		&+\tr( (\Xm- \Xmat_i^m) \sigLapm_o (\Xm- \Xmat_i^{m})^{T})\bigg)
		+c
	\end{split}
\label{eq:init_obj10}
\end{equation}
where $c$ stands for a constant. Here the matrix $ \Xmat_i^{m}\in \R^{J N^m \times K^m}$ has $\xim$ in its $i$-th column, and its other columns are $0$. The gradient of the first term in \eqref{eq:init_obj10} contains the expression $( \Xmat_i^m) \sigLapm_d$, which gives the $i$-th column $ \xim$ of $\Xmat_i^m$ since $\sigLapm_d=\eye_{K^m} \in \R^{K^m \times K^m}$ is the identity matrix. Defining the matrices
\begin{equation}
\begin{split}
\widetilde{\Amat}=\eta_{c} \, \eye_{JN^m},
\qquad
\widetilde{\Bmat}=2\eta_{c} \, (\Xm- \Xmat_i^m) \sigLapm_o,
\end{split}
\label{eq:init_obj11}
\end{equation}
and denoting the $i$-th column of $\widetilde \Bmat$ as $\widetilde \bvect_i$, we get the optimal $\xvect^{k+1}$ vector in the problem \eqref{eq_obj8_xupd} by solving 
\begin{equation}	
	\begin{split}
		&2(\Amat+\widetilde{\Amat}) \xvect^{k+1} + (\bvect+\widetilde{\bvect_i})=0.\\
	\end{split}
\label{eq:init_obj12}
\end{equation}
Finally, the optimal  $\zvect^{k+1}$ vector in the problem \eqref{eq_obj8_zupd} can simply be found with the shrinkage operation \cite{BoydPCPE11}.

We continue the iterations \eqref{eq_obj8_xupd}-\eqref{eq_obj8_uupd} until the convergence of the ADMM algorithm, which yields the optimal $\xvect= \xim$ vector. Repeating this operation iteratively for each $\xim$ vector, we obtain the optimized coefficient matrix $\{ \Xm\}$ for each graph.

\subsubsection{Optimization of the Kernel Parameters $\psivect$}
\label{ssec:OptPhi}

Fixing the coefficient vectors $\{\Xm\}$, the optimization problem  \eqref{eq_obj_final_reg_map} becomes
\begin{equation}	
	\begin{split}
		&\min\limits_{\psivect} f(\psivect)
	=\min\limits_{\psivect}\sum_{j=1}^{J}(\mu_j)^2+\eta_{s}\sum_{j=1}^{J}(s_j-s_0)^2\\
	&+\eta_{w}\sum_{m=1}^{M}\sum_{i=1}^{K^m}\norm{ \Sim \yim- \Sim \Dmpsi \xim}^2\\
		&+\eta_{y}\sum_{m=1}^{M} \tr(( \Xm)^{T}(\Dmpsi)^{T} \Lapm \Dmpsi \Xm).
	\end{split}
\label{eq_init_obj13}
\end{equation}
While the objective function in \eqref{eq_init_obj13} is not convex in $\psivect$, it is a differentiable function. We thus seek a local minimum of this function with the gradient descent algorithm. The computation of the gradient of the objective function is explained in Appendix A.

\subsubsection{Overall Algorithm}

Since the objective function  in \eqref{eq_obj_final_reg_map} is nonnegative and it remains non-increasing during the iterative updates discussed in Sections \ref{ssec:OptXm} and \ref{ssec:OptPhi}, it is guaranteed to converge. We repeat the iterations until the convergence of the objective function, and thus obtain the kernel parameters $\psivect$ and the coefficients $\{ \Xm \}$. We next explain the two possible utilization modes of the learnt model.

\textit{Transductive setting.} The partially known signals on each graph $\mathcal{G}^m$ can be fully reconstructed as
$
\Ym  = \Dmpsi \Xm 
$
and the initially missing observations are obtained by retrieving the corresponding entries of the matrix $\Ym$.

\textit{Inductive setting.} Our method can also be employed in an inductive setting, where the learnt model is used to reconstruct a new set of test signals $\Ymat_t^m$ on each graph $\Gm$ that were not available in the training. The coefficients $\Xmat_t^m$ of $\Ymat_t^m$  based on the learnt kernels $\psivect$ can then be computed by solving the problem \eqref{eq:init_obj4} as described in Section \ref{ssec:OptXm}, where the selection matrices are formed according to the missing entries of $\Ymat_t^m$ and the signal graph Laplacian $\sigLapm $ is updated by expanding $\sigGm $ with the new test data $\Ymat_t^m$. The test signals are finally reconstructed as $\Ymat_t^m  = \Dmpsi \Xmat_t^m$.

We call the proposed method Spectral Graph Kernel Learning (SGKL) and summarize it in Algorithm \ref{alg_SGKL}. In order to investigate the computational complexity of our method, let us assume $\Nm=N$ and $\Km = K$ for all $m$ for simplicity. The complexity of computing the sparse coefficients with the ADMM method is of $O(MKJ^3N^3)$, while the optimization of dictionary parameters via gradient descent is of complexity $O(MKJ^2N^3+MK^2N)$. Therefore, the overall complexity of the proposed algorithm is obtained as $O(MKJ^3N^3+MK^2N)$.

\begin{algorithm}[h]
\footnotesize
\caption{Spectral Graph Kernel Learning (SGKL) }
\begin{algorithmic}[1]

\STATE
\textbf{Input: } Partially available signals $\{\Ym\}$, graphs $\{\Gm\}$

\STATE
\textbf{Initialization: } 

Form graph Laplacians $\{ \Lapm\}$ and signal Laplacians $\{\sigLapm \}$.

Set $\psivect$ randomly.

Find coefficients $\{\Xm \}$ by solving \eqref{eq:init_obj5}.

\STATE
\textbf{until} Convergence of objective function

\STATE \hspace{\algorithmicindent}
\label{alg_sgkl_ker_upd}
Find kernel parameters $\psivect$ by solving \eqref{eq_init_obj13}.

\STATE \hspace{\algorithmicindent}
\label{alg_sgkl_coef_upd}
Find coefficients $\{\Xm \}$ by solving \eqref{eq:init_obj5}.

\STATE
\textbf{end}

\STATE
\textbf{Output: } 

Learnt kernel parameters $\psivect$, narrowband signal prototypes $\Dmpsi $
Reconstructed signals $\Ym = \Dmpsi \Xm $ \\

\end{algorithmic}
\label{alg_SGKL}
\end{algorithm}
\normalsize
%

%

\section{Theoretical Analysis of the Proposed Method}
\label{sec_theo_anlys}

We now present a theoretical performance analysis of learning narrowband graph kernels from incomplete data, where we aim to study the estimation accuracy of the learnt representation with respect to factors such as the number of signals, number of graphs, noise level, and sparsity of the signals. A particular question we would like to answer is when learning a model jointly on multiple graphs is preferable to learning individual models on single graphs. Although in Section \ref{ssec_graph_signal_model} we motivated our algorithm by assuming that the signals $\{\yim\}$ on different graphs $\G^1, \dots, \G^M$ admit representations of the form $\yim = \Dmpsi \xim + \wim$ through a common parameter vector $\psivect$ in the spectral domain, in real applications there is likely to be some deviation between the spectra of signals acquired on different, independently constructed graphs. We then wish to characterize the trade-off between the positive effect of an enlarged data size due to the incorporation of multiple graphs in the learning, and the negative effect of the mismatch between the parameter vectors $\psivect$ of the signals on different graphs.

For the tractability of the derivations, we focus on a setting where no regularization is made on the smoothness of graph signals and the coefficients, i.e., we consider that $\eta_y = \eta_c =0$ in \eqref{eq_obj_final_reg_map}. In order to study the effect of a possible discrepancy between the spectra of the signals on different graphs, we consider a signal model
\begin{equation}
\label{eq_new_sig_model}
\yim = \Dm (\psim) \xim + \wim
\end{equation}
 which extends the one in Section \ref{ssec_graph_signal_model}, such that the spectral parameters $\psivect^1, \dots, \psivect ^M$ generating the signals on different graphs may differ from each other. We consider that a single common estimate $\psiest$ is learnt for the parameter vectors $\psim$ on different graphs. Staying in line with the iterative optimization algorithm described in Algorithm \ref{alg_SGKL}, let $\ximestt$ denote the  estimates of the coefficients $\xim$ at some iteration $t$ of the algorithm. Let $t=T$ denote the last iteration of the algorithm. Then, the common estimate $\psiest$ of the parameter vector in Step \ref{alg_sgkl_ker_upd} of the last iteration is found by solving
\begin{equation}
\label{eq_defn_psiest}
\begin{split}
\psiest = \arg \min_{\psivect \in \Psispace} \ & \sum_{j=1}^{J}(\mu_j)^2+\eta_{s}\sum_{j=1}^{J}(s_j-s_0)^2\\
	&+\eta_{w}\sum_{m=1}^{M}\sum_{i=1}^{\Km}\norm{ \Sim \yim- \Sim \Dmpsi \ximestLast }^2
\end{split}
\end{equation}
where $\Psispace \subset \R^{2J}$ is assumed to be a compact parameter domain that is void of zero scale parameters. Consequently, in Step \ref{alg_sgkl_coef_upd} of the last iteration, the corresponding estimate $\ximest(\psiest)$ of the coefficient vector $\xim$ of each signal $\yim$ is obtained as 
\begin{equation}
\label{eq_defn_ximest}
\begin{split}
\ximest(\psiest) = \arg \min_{\xvect \in \R^{J \Nm}}  \ \eta_{x}  \norm{\xvect }_1 +\eta_{w} \norm{ \Sim \yim- \Sim \Dmpsiest \xvect}^2.
\end{split}
\end{equation}
which constitutes the final output of the algorithm.

In our analysis, we model the selection matrices $\Sim $ determining the observed entries $\Sim \yim$ of the signals as random, i.i.d.~matrices, which are also independent of $\yim$, $\xim$, and $\wim$. Let $\ym=\Dmpsim \xm + \wm$ represent a signal on graph $\Gm$ with coefficients $\xm$ and noise component $\wm$, sampled from the same distribution as the signals $\{\yim\}$ in the data set, such that the random vectors $\{\xm\} \cup \{\xim\}_{i=1}^\Km  $ are i.i.d., as well as $\{\wm\} \cup \{\wim\}_{i=1}^\Km  $  and $\{\ym\} \cup \{\yim\}_{i=1}^\Km  $. Similarly, let $\Sm$ denote a selection matrix having an independent and identical distribution as that of $\{\Sim\}$. Let us also define the complement $\Smc \in \mathbb{R}^{(N^m-R^m) \times N^m}$ of a selection matrix $\Sm \in \mathbb{R}^{R^m \times N^m}$ as the matrix that selects the unobserved entries of a signal $\ym$ through the product $\Smc \ym$, where $\Rm$ and $\Nm - \Rm$ respectively denote the number of nodes where the signal $\ym$ is observed and unobserved. 

For any fixed parameter vector $\psivect_0 \in \Psispace$, let 
 \begin{equation}
 \begin{split}
 \label{eq_defn_xmest_psi0}
 \xmest(\psivect_0) &= \arg \min_{\xvect \in \R^{J \Nm}}  \ \eta_{x}  \norm{\xvect }_1 \\
  &+\eta_{w} \norm{ \Sm \ym- \Sm \Dm(\psivect_0) \xvect}^2
\end{split}
 \end{equation}
denote the estimate of the coefficient vector of the signal $\ym$ on $\Dm(\psivect_0)$. Let us also define the errors
\begin{equation*}
\begin{split}
\errtrainm(\psivect_0)&= \frac{1}{\Rm} \| \Sm \ym - \Sm \Dm(\psivect_0) \xmest(\psivect_0) \| \\
\errtrainim(\psivect_0)&= \frac{1}{\Rim} \| \Sim \yim - \Sim \Dm(\psivect_0) \ximest(\psivect_0) \| \\
\end{split}
\end{equation*}
which respectively denote the average approximation errors of the observed parts of the signals $\ym$ and $\yim$ when represented in the dictionary  $\Dm(\psivect_0)$. Similarly let
\begin{equation*}
\begin{split}
\errtestm(\psivect_0)&= \frac{1}{\Nm - \Rm} \| \Smc \ym - \Smc \Dm(\psivect_0) \xmest(\psivect_0) \| \\
\errtestim(\psivect_0)&= \frac{1}{\Nm - \Rim} \| \Simc \yim - \Simc \Dm(\psivect_0) \ximest(\psivect_0) \| \\
\end{split}
\end{equation*}
denote the average estimation errors of the unobserved components of the same graph signals.  

While we maintain a generic treatment without any assumptions about the particular distribution of $\Sim$ matrices, we make a couple of mild assumptions in our problem setting as follows. We first trivially assume that the signals $\yim, \ym$ are observed on at least one graph node, i.e., $\Rim, \Rm \geq 1$.  Next, we assume that the estimate $\psiest$ in \eqref{eq_defn_psiest} satisfies
\begin{equation}
\label{eq_assum_errdata_smaller}
\begin{split}
\frac{1}{M} \sum_{m=1}^M \frac{1}{\Km} \sum_{i=1}^\Km E[\errtestim(\psiest)] 
\leq
\frac{1}{M} \sum_{m=1}^M E[\errtestm(\psiest)]
\end{split}
\end{equation}
and also that there exists a constant $\csmooth>0$ such that for any parameter vector $\psivect_0 \in \Psispace $  
\begin{equation}
\label{eq_assum_csmoothm}
\begin{split}
 \quad & \big | E\left[ \errtestm(\psivect_0) \right]  - E\left[ \errtrainm(\psivect_0) \right]  \big | \leq \csmooth
\end{split}
\end{equation}
for all $m=1, \dots, M$. The condition \eqref{eq_assum_errdata_smaller}  is a very mild assumption ensuring that the expected error of the unobserved component $\Simc \yim$ of the graph signals in the data set used for learning $\psiest$ do not exceed that of a new graph signal $\ym$ drawn independently from the same distribution but not used in learning $\psiest$.  Then, the condition \eqref{eq_assum_csmoothm} imposes that the graph signals $\ym$ have a sufficiently regular variation on the topologies of the graphs $\Gm$ and the sampling defined by the distribution of the matrices $\Sm$ be sufficiently uniform, so that the expected average representation error of the unobserved component $\Smc \ym$ of the signal $\ym$ does not deviate from that of its observed component $\Sm \ym$ by more than $\csmooth $ for a representation computed over an arbitrary dictionary $\Dm(\psivect_0)$. The upper bound $\csmooth $ is kept as a generic problem parameter in our analysis and would depend on several practical factors such as the topology of the graphs $\Gm$, the  distribution of the selection matrices $\Sim$, $\Sm $, and the parameter domain $\Psispace $. 

Finally, let us define the following distance $\dist$ between $\psione, \psitwo \in \Psispace $
\begin{equation}
\label{eq_defn_distance_d}
\begin{split}
\dist(\psione, \psitwo) \triangleq \max_{m=1, \dots, M} E \left[ \ |  \errtrainm(\psione) - \errtrainm(\psitwo) | \ \right].
\end{split}
\end{equation}
It is easy to verify that $\dist $ is a pseudometric on $\Psispace $. In our analysis, we consider the parameter domain $\Psispace $ to be compact with respect to the pseudometric $\dist $. Hence, for any $\epsilon>0$, one can find a finite set $\covcents=\{\psivect_l \}_{l=1}^\covnumeps \subset \Psispace $ of parameter vectors, such that for any $\psivect_0 \in \Psispace$, there exists some $\psivect_l \in \covcents$ with $\dist(\psivect_0, \psivect_l) < \epsilon$. The minimal value of $\covnumeps$ with this property is then called the covering number of $\Psispace $ with radius $\epsilon$.

%

We now present our first main result in the following statement. 

\begin{theorem}
\label{thm_dev_unobs_obs}
Let $0<\cprob <1$ and $0<\varepsilon <1$ be any constants and let $\cparm \triangleq E[(\Rm)^{-2}] \, (2 \Nm \sigma^2 + 4 (J \Nm \delta)^2 + \eta_x \eta_w^{-1} J \Nm \delta )  $. Then with probability at least
\begin{equation}
\label{eq_thm1_prob_expr}
1- \cprob - \frac{\covnumcprobeps  \ \sum_{m=1}^M \frac{\cparm}{\Km} }{ M^2 \, \varepsilon^2},
\end{equation}
the expected average estimation error of the unobserved samples $\Simc \yim$ of the graph signals $\yim$ given by Algorithm \ref{alg_SGKL} is upper bounded as
\begin{equation*}
\begin{split}
E \left[ \frac{1}{M} \sum_{m=1}^M  \frac{1}{\Km} \sum_{i=1}^\Km \errtestim(\psiest) \right] 
\leq 
 \frac{1}{M} \sum_{m=1}^M \frac{1}{\Km} \sum_{i=1}^\Km \errtrainim(\psiest) \\
  + \csmooth + (2+\cprob) \varepsilon.
 \end{split}
 \end{equation*}
\end{theorem}

The proof of Theorem \ref{thm_dev_unobs_obs} is given in Appendix B.  Algorithm \ref{alg_SGKL}  computes a solution $\psiest$ that fits to the observed components $\Sim \yim$ of the signals by minimizing their approximation error; however, a critical question is how well the model given by $\psiest$ generalizes to the unobserved components $\Simc \yim$ of the same signals.  Theorem  \ref{thm_dev_unobs_obs} then states that the expected estimation error of the unobserved components $\Simc \yim$ will not deviate from the empirically found approximation error of the observed components $\Sim \yim$ by more than an amount $ \csmooth + (2+\cprob) \varepsilon$ with the probability guaranteed in \eqref{eq_thm1_prob_expr}. In order to better interpret the statement of the theorem, let us contemplate a simple setting where the number of signals are of comparable order across different graphs so that $\Km = \Theta( \Kc)$, meaning that $\Km$ is proportional to some $\Kc$ for all graphs $m=1, \dots, M$. Let us also denote $\cpar \triangleq \max_{m=1, \dots, M} \cparm$. Then, the theorem states that the estimates of the unobserved signal components attain the accuracy level determined by $\varepsilon$ and $\tau$  with probability 
\begin{equation*}
\begin{split}
1- \cprob - \frac{\covnumcprobeps \, \cpar }{ M K \, \varepsilon^2} 
= 1- \cprob - O \left( \frac{\cpar}{ M K} \right).
\end{split}
\end{equation*}
Then letting $\cprob \ll 1$, the probability of attaining some target estimation accuracy improves at rate $1- O(1/ (MK))$ as the number of graphs $M$ and the number of signals $K$ observed over each graph increases. Meanwhile, the dependence of the constant $\cpar=O(\sigma^2 + \delta^2)$ on the model parameters $\sigma$ and $\delta$ suggests that the estimation performance is negatively affected by the increase in the signal noise level $\sigma^2$, pointing to the fact that data exhibiting larger deviation from the presumed model is more difficult to predict. The estimation performance also degrades as the Laplace distribution parameter $\delta$ of the signal coefficients $\xim$ grows. Larger values of $\delta$ are associated with less sparse signal representations and thus a signal model of higher complexity, which would necessitate more data samples (e.g. so as to ensure $MK=O(\delta^2)$) for properly learning a signal model. Due to the relation $\cparm = O(E[(\Nm/\Rm)^2])$, the estimation performance is also expected to improve at an inverse quadratic rate as the number $\Rm$ of available signal components increases.

While Theorem \ref{thm_dev_unobs_obs} characterizes the estimation error of unobserved signal components in terms of the approximation error of the observed ones, this raises the question of how much the algorithm may be expected to decrease the approximation error of the observed signal components; i.e., how well the computed model $\psiest$ fits to the available signals at hand. This depends on how well the actual data adheres to the assumed model. In particular, in a realistic scenario there would be some deviation between the spectral contents of signals acquired on different graphs. In what follows, we study how such a discrepancy affects the learning performance. We consider a scenario where the signals $\yim = \Dm (\psim) \xim + \wim$ on each graph stem from a different parameter vector $\psim \in \Psispace $ with
\begin{equation}
\label{eq_dev_psim_multigraph}
\begin{split}
\psim = \psicom + \psidevm.
\end{split}
\end{equation}
Here $\psicom \in \Psispace $ represents a common parameter vector and $\psidevm $ represents the deviation of the spectra of the signals  on each graph from this common spectrum. We assume that this deviation is bounded such that $\| \psidevm \| \leq \delpsi$ for some spectrum discrepancy parameter $\delpsi>0$, for all $m=1, \dots, M$. Our purpose is then to study how the spectrum discrepancy $\delpsi$ affects the algorithm performance. 




For simplicity, let us consider that the number of signals $\Km= \Theta( \Kc)$, total nodes $\Nm= \Theta( \Nc)$ and observed nodes $\Rim = \Theta(\Rc)$ take comparable values across different graphs, the variables $\Kc$, $\Nc$ and $\Rc$ standing for their orders of magnitude.  Then in the following result, we present an upper bound on the expected approximation error of the observed signal components. 
\begin{theorem}
\label{thm_exp_emp_err}
In the multi-graph setting described by \eqref{eq_dev_psim_multigraph}, assume that the solution given by Steps \ref{alg_sgkl_ker_upd} and \ref{alg_sgkl_coef_upd} of Algorithm \ref{alg_SGKL} converges, i.e., $\ximest(\psiest) = \ximestLast$ in  \eqref{eq_defn_psiest}, \eqref{eq_defn_ximest}.  Then, there exist constants $\cx, \cD, \cf $ such that the expected approximation error of the observed signal components is bounded as
\begin{equation*}
\begin{split}
& E\left[ \frac{1}{M} \sum_{m=1}^M \frac{1}{\Km} \sum_{i=1}^\Km \errtrainim(\psiest) \right] 
 \leq
 \frac{\sqrt{2 \Theta(\Rc)} \sigma  + 2 \cx  \sqrt{J \Theta(N)}}{\Theta(\Rc)} \\
   &+
  \frac{\cf}{\sqrt{\eta_w M \Theta(\Kc)} \Theta(\Rc)} 
  +\frac{2 \cx \cD \delpsi \sqrt{J \Theta(N)}}{\Theta(\Rc)}.
\end{split}
\end{equation*}
\end{theorem}
The proof of Theorem \ref{thm_exp_emp_err} is given in Appendix C. The theorem puts forward how different components of the expected approximation error vary with the problem parameters. The first error component is determined by the noise variance $\sigma^2$ and the sparsity parameter $\delta$ through the term $\cx$. This component gets smaller as the model complexity and the deviation of the signals from the model decrease (i.e., small $\delta$ and $\sigma$), independently of $\Kc$ and $\delpsi$. It is particularly interesting to examine the second and the third error components. The second component decays at rate $O(1/\sqrt{M \Kc})$ as the number $M$ of graphs and the number $\Kc$ of graph signals increases, thus characterizing the improvement in the accuracy with the amount of data. Meanwhile, the third component shows how the accuracy is affected from the spectral discrepancy $\delpsi$, the error increasing at a linear rate of $O(\delpsi)$ as $\delpsi$ increases.

Combining the results in Theorems \ref{thm_dev_unobs_obs} and \ref{thm_exp_emp_err}, together with a simple application of Markov's inequality, we arrive at the following corollary.
\begin{corollary}
\label{cor_exp_test_error}
Let constants $0<\cprob <1$,  $0<\varepsilon <1$, and $\ctesterr>0$. With probability at least
\begin{equation}
\label{eq_cor_exp_test_error}
\begin{split}
&1- \cprob - \frac{\covnumcprobeps  \ \sum_{m=1}^M \frac{\cparm}{\Km} }{ M^2 \, \varepsilon^2} 
- 
\frac{1}{\ctesterr} \bigg(
\frac{\sqrt{2 \Theta(\Rc)} \sigma  + 2 \cx  \sqrt{J \Theta(N)}}{\Theta(\Rc)} \\
 & +
 \frac{\cf}{\sqrt{\eta_w M \Theta(\Kc)} \Theta(\Rc)}  
 +
  \frac{2 \cx \cD \delpsi \sqrt{J \Theta(N)}}{\Theta(\Rc)}
  \bigg)
\end{split}
\end{equation}
the expected estimation error of unobserved samples is upper bounded as
\begin{equation}
\label{eq_cor_testerr_acc_bound}
\begin{split}
&E \left[ \frac{1}{M} \sum_{m=1}^M  \frac{1}{\Km} \sum_{i=1}^\Km \errtestim(\psiest) \right] 
\leq 
 \frac{1}{M} \sum_{m=1}^M \frac{1}{\Km} \sum_{i=1}^\Km \errtrainim(\psiest) \\
 & + \csmooth + (2+\cprob) \varepsilon
\quad  \leq  \quad 
\ctesterr + \csmooth + (2+\cprob) \varepsilon.
\end{split}
\end{equation}
\end{corollary}

\textbf{Remark 1.} Corollary \ref{cor_exp_test_error} provides some insight on when joint learning on multiple graphs achieves better estimation accuracy than individual learning on single graphs. This decision essentially depends on how large the spectral discrepancy $\delpsi$ is compared to the number $\Kc$ of graph signals, as a result of the trade-off between the decay rate of $O(1/\sqrt{M \Kc})$ and the error term of $O(\delpsi)$ in \eqref{eq_cor_exp_test_error}. In Appendix G, we use Corollary \ref{cor_exp_test_error} to show that when the amount of data is limited to
\begin{equation}
\label{eq_cond_joint_over_single_learn}
\begin{split}
\Kc < O\left(\frac{1}{\delpsisq}\right),
\end{split}
\end{equation}
then jointly learning a common model $\{ \psiest \}, \{\ximest (\psiest)\}_{m=1}^M$ using the signals on all graphs $\G^1, \G^2, \dots, \G^M$ results in better estimation performance than learning an individual model  $\{ \psiest^m \}, \{\ximest (\psiest^m)\}$ on each graph $\Gm$. This can be interpreted in the way that, when the data sets on all graphs bear sufficiently high similarity in terms of their spectral content, the lack of data on the individual graphs can be compensated by combining the data sets from all graphs and learning a common model. On the other hand, when the spectral discrepancy $\delpsi$ is too high, or there are considerably many signals available on each single graph $\Gm$ (such that $\delpsi > O(1/\sqrt{\Kc})$ as opposed to the condition \eqref{eq_cond_joint_over_single_learn}), then it may be preferable to learn an individual and independent model on each graph. The relation  \eqref{eq_cond_joint_over_single_learn} shows that the tolerable spectral discrepancy to advocate joint learning over individual learning varies at an inverse square root rate $\delpsi = O(1/\sqrt{\Kc})$ as the data size $\Kc$ grows.


\section{Experimental Results}
\label{sec_exp_results}

In this section, we evaluate the performance of our method with experimental results. In Section  \ref{sec:sensitivity_analysis}, we present a sensitivity analysis of our method and in Section \ref{compExp} we conduct comparative experiments. We use the following data sets in our experiments:




\textit{1. Synthetic data set.} We construct two independent graphs $\mathcal{G}^1$ and $\mathcal{G}^2$ with $N^1 = N^2 = 100$ nodes, with a $10$-NN topology and uniformly distributed node locations. A graph dictionary is generated on each graph with $J=4$ kernels, where the ground truth kernel parameters $\mu_j, s_j$ are randomly sampled from normal distributions and are kept fixed throughout the experiments. $K^1=200$ and $K^2=400$ graph signals $\{ \yvect_i^1 \}, \{ \yvect_i^2 \}$  are generated on  $\mathcal{G}^1$ and $\mathcal{G}^2$ following the model \eqref{eq_signal_model}.





\textit{2. Mol\`{e}ne data set.} The Mol\`{e}ne data set consists of meteorological measurements taken at various weather stations in the Brittany region of France  \cite{Girault-Stationarity}. We construct two graphs $\mathcal{G}^1$ and $\mathcal{G}^2$ with a $5$-NN topology from $N^1=37$ and $N^2=31$ stations collecting temperature and wind speed measurements, respectively. We experiment on $K^1=K^2=744$ temperature and wind speed signals.



%

\textit{3. COVID-19 data set.} The COVID-19 data set consists of the number of daily new
cases recorded in $N^1 = 37$ European countries and $N^2 = 50$ USA states  \cite{COVID-cases}. The graphs  $G^1$ and $G^2$ are constructed with $4$-NN and $3$-NN topologies based on a hybrid distance measure combining geographical proximities and flight frequencies \cite{COVID-flights}. The experiments are conducted on $K^1 = K^2 = 483$ graph signals representing population-normalized daily new case numbers. 


\textit{4. NOAA data set.}  The NOAA weather data set contains hourly temperature measurements for one year taken in weather stations across the United States averaged over the years 1981-2010 \cite{NOAA-dataset}. We construct a  $7$-NN graph from $N = 246$ weather stations  with Gaussian edge weights. The experiments are done on $8760$ graph signals.

\subsection{Performance and Sensitivity Analysis}\label{sec:sensitivity_analysis}


We first analyze the sensitivity of our method to noise, number of kernels, and algorithm weight parameters using the synthetic data set generated with ground truth sparsity  $40$ and a missing observation ratio of $20    \%$. The missing observations are estimated with the proposed SGKL algorithm and the normalized mean square error (NMSE) of the estimates is computed as
\begin{equation}
NMSE^m = \frac{\norm{\unkyvect^m - \tilde{\unkyvect}^m}^2}{\norm{\unkyvect^m}^2}
\label{eq:NMSE}
\end{equation}
and averaged over different realizations of the experiments with random selections of the missing observations, where $\unkyvect^m$ is the vector obtained by concatenating the missing entries of the signals $\yvect_{1}^m, \dots, \yvect_{{K^m}}^m$, and  $\tilde{\unkyvect}^m$ is its estimate. 



\subsubsection{Effect of noise level} \label{SNR}

In this first experiment, we generate graph signal sets with different noise levels and examine the effect of noise on the algorithm performance. The NMSE of the proposed SGKL method is reported with respect to the SNR of the graph signals in Table \ref{table:1}.  The error decreases monotonically with increasing SNR as expected, dropping below 0.1 for SNR values above $6$ dB. The algorithm performance seems to be rather robust to noise, given that the NMSE remains around $0.4$ even at $-1$ dB SNR, where the signal power is less than the noise power.

\begin{table}[t!]
\centering
\begin{tabular}{||c | c | c ||} 
 \hline
 SNR, dB & $\mathcal{G}^1$ & $\mathcal{G}^2$ \\ [0.5ex] 
 \hline\hline
 -8 & 1.61 & 1.49  \\
 \hline
 -6 & 1.38 & 1.31  \\
  \hline
 -5 & 1.02 & 1.02  \\
  \hline
 -3 & 0.61 & 0.59  \\
  \hline
 -1 & 0.43 & 0.42  \\
  \hline
  6 & 0.13 & 0.12  \\
  \hline
  9 & 0.01 & 0.01  \\
  \hline
  15 & 0.003 & 0.004  \\
  \hline
  21 & $4\cdot10^{-4}$ & $5\cdot10^{-4}$  \\
  \hline
  35 & $4\cdot10^{-4}$ & $4\cdot10^{-4}$  \\
 \hline
 \end{tabular}
\caption{NMSE vs SNR for $\mathcal{G}^1$ and $\mathcal{G}^2$}
\label{table:1}
\end{table}


\subsubsection{Effect of number of kernels} \label{KernelN}

We next study the effect of model mismatch through the number of graph kernels generating the signals. We generate the ground truth graph signals with $J_{GT} = 4$ kernels and learn dictionary models with a varying number of kernels $J$. The variation of the NMSE with the choice of $J$ is shown in Table \ref{table:2}. Even in case of a mismatch between the number of learnt kernels $J$ and the  number of ground truth kernels $J_{GT}$, the proposed SGKL algorithm is still able to attain estimation errors comparable to the ideal case $J=J_{GT}=4$. In fact, we have observed that for $J < J_{GT}$ the algorithm tends to learn kernels with wider spectra in order to compensate for the lack of spectral content, while for $J > J_{GT}$, it may tend to center the extra spectral kernel at high frequencies, which may increase the error due to too high vertex concentration.  Surprisingly, selecting $J = 3$ yields lower NMSE than the ideal case $J=4$, which can be explained via the principle that reducing model complexity might improve generalization capability in machine learning.


\begin{table}[t!]
\centering
\begin{tabular}{||c | c | c ||} 
 \hline
 $J$ & $\mathcal{G}^1$ & $\mathcal{G}^2$ \\ [0.5ex] 
 \hline\hline
 1 & 0.0027 & 0.0016  \\ 
 \hline
 2 & $4.38 \cdot10^{-4}$ & $3.71\cdot10^{-4}$  \\ 
  \hline
 3 & $3.36\cdot10^{-4}$ & $2.53\cdot10^{-4}$  \\
  \hline
 4 & $4.06\cdot10^{-4}$ & $3.6\cdot10^{-4}$  \\
  \hline
 5 & 0.0039 & 0.0051  \\
 \hline
 \end{tabular}
\caption{NMSE vs $J$ for $\mathcal{G}^1$ and $\mathcal{G}^2$}
\label{table:2}
\end{table}

\subsubsection{Sensitivity to hyperparameter selection} \label{Sensitivity}

We next examine the sensitivity of the algorithm performance to the choice of the weight parameters $\eta_s$, $\eta_x$, $\eta_w$, $\eta_y$, and $\eta_c$. In each experiment, we change only one of weights while fixing the others. The variation of the NMSE with respect to the weight parameters is presented in Tables \ref{table:5}-\ref{table:9}.

\begin{table}[t!]
\centering
\begin{tabular}{||c | c | c ||} 
 \hline
 $\eta_s$ & $\mathcal{G}^1$ & $\mathcal{G}^2$ \\ [0.5ex] 
 \hline\hline
 $10^5 $& 0.12 & 0.14  \\ 
 \hline
 $10^6$ & 0.027 & 0.028  \\ 
  \hline
 $10^7 $& $4.14\cdot10^{-4} $& $2.49\cdot10^{-4} $ \\
  \hline
 $10^8$ &$ 4.06\cdot10^{-4}$ & $3.6\cdot10^{-4} $ \\
  \hline
 $10^9$ & $8.2\cdot10^{-4} $&$ 8.4\cdot10^{-3} $ \\
  \hline
 $10^{10}$ & 0.0013 & 0.0013  \\ 
 \hline
 \end{tabular}
\caption{NMSE vs $\eta_s$ for $\mathcal{G}^1$ and $\mathcal{G}^2$}
\label{table:5}
\end{table}


Recalling that the parameter $\eta_s$ controls the deviation of the kernel scale parameters from their nominal values, from Table \ref{table:5}, we can conclude the effectiveness of learning spectral kernels with close-to-nominal bandwidth values. Even when $\eta_s$ is increased beyond its optimal value, the proposed SGKL method still provides NMSE values in the order of $10^{-3}$, efficiently adapting the representation coefficients to the presumably suboptimal choice of the kernels. The results in Table \ref{table:6} show that, when the sparsity-controlling $\eta_x$ parameter is selected too high, the algorithm is forced to represent graph signals with a smaller number of dictionary atoms, resulting in a higher NMSE value.  In Table \ref{table:7} the optimal value of the weight $\eta_w$ of the data fidelity term is seen to be around $10^6$. While setting  $\eta_w$ to too low values impairs the learning, the performance is rather robust to overly high values of $\eta_w$, the extent of overfitting remaining tolerable. The optimal value of the signal smoothness weight $\eta_y$ is seen to be around $10^2$ in Table \ref{table:8}. Although too high values of $\eta_y$ lead to the oversmoothing of the reconstructed signals, the algorithm performance is stable over a relatively wide range of $\eta_y$ values. Lastly, the results in Table \ref{table:9} show that learning graph signal representations by exploiting the information of the available entries of similar signals reduces the estimation error significantly. Although excessively small choices of $\eta_c$ weakens the capability of learning from cross entries and increases the estimation error, the algorithm performance is seen to be generally robust to the choice of $\eta_c$.




\begin{table}[t!]
\centering
\begin{tabular}{||c | c | c ||} 
 \hline
 $\eta_x$ & $\mathcal{G}^1$ & $\mathcal{G}^2$ \\ [0.5ex] 
 \hline\hline
 5 & $7.3\cdot10^{-4}$ &$ 5.56\cdot10^{-4}$  \\ 
 \hline
 50 &$ 2.02\cdot10^{-4}$ & $2.13\cdot10^{-4} $ \\ 
  \hline
 500 &$ 4.06\cdot10^{-4}$ & $3.6\cdot10^{-4} $ \\
  \hline
  $5\cdot10^3$ & 0.0055 & 0.0058  \\
  \hline
$ 5\cdot10^4$ & 0.22 & 0.21  \\
  \hline
 $5\cdot10^5$ & 1 & 1  \\
  \hline
 $5\cdot10^6$ & 1 & 1  \\ 
 \hline
 \end{tabular}
\caption{NMSE vs $\eta_x$ for $\mathcal{G}^1$ and $\mathcal{G}^2$}
\label{table:6}
\end{table}

\begin{table}[t!]
\centering
\begin{tabular}{||c | c | c ||} 
 \hline
 $\eta_{\omega}$ & $\mathcal{G}^1$ & $\mathcal{G}^2$ \\ [0.5ex] 
 \hline\hline
 $10^3$ & 0.23 & 0.23  \\ 
  \hline
 $10^4$ & 0.003 & 0.003  \\
  \hline
 $ 10^5$ & $4.06\cdot10^{-4}$ & $3.6\cdot10^{-4} $ \\
  \hline
 $10^6$ & $3.88\cdot10^{-4} $& $3.79\cdot10^{-4} $ \\
  \hline
 $10^7$ & $4.84\cdot10^{-3}$ & $4.76\cdot10^{-3} $ \\
  \hline
 $10^8$ & $9.7\cdot10^{-3}$ &$ 1.1\cdot10^{-2} $ \\ 
 \hline
 \end{tabular}
\caption{NMSE vs $\eta_w$ for $\mathcal{G}^1$ and $\mathcal{G}^2$}
\label{table:7}
\end{table}


\begin{table}[t!]
\centering
\begin{tabular}{||c | c | c ||} 
 \hline
 $\eta_{y}$ & $\mathcal{G}^1$ & $\mathcal{G}^2$ \\ [0.5ex] 
 \hline\hline
 $10^0$ & 0.0042 & 0.0052  \\ 
 \hline
$ 10^1$ & 0.0091 & 0.010  \\ 
  \hline
 $10^2$ & $4.06\cdot 10^{-4}$  & $3.6\cdot10^{-4}$  \\
  \hline
  $10^3$ & 0.0034 & 0.041  \\
  \hline
 $10^4 $& 0.013 & 0.014  \\
  \hline
 $10^5 $& 0.022 & 0.022  \\
  \hline
 \end{tabular}
\caption{NMSE vs $\eta_y$ for $\mathcal{G}^1$ and $\mathcal{G}^2$}
\label{table:8}
\end{table}

\begin{table}[t!]
\centering
\begin{tabular}{||c | c | c ||} 
 \hline
 $\eta_{c}$ & $\mathcal{G}^1$ & $\mathcal{G}^2$ \\ [0.5ex] 
 \hline\hline
 $10^0$ & 0.01 & 0.01  \\ 
 \hline
 $10^1 $ & 0.006 & 0.006  \\ 
  \hline
 $10^2$ & 0.002 & 0.002  \\
  \hline
  $10^3 $& $4.06\cdot 10^{-4} $ & $3.6\cdot10^{-4} $ \\
  \hline
 $10^4$ & 0.003 & 0.002  \\
  \hline
 $10^5$ & 0.003 & 0.003  \\
  \hline
 \end{tabular}
\caption{NMSE vs $\eta_c$ for $\mathcal{G}^1$ and $\mathcal{G}^2$}
\label{table:9}
\end{table}

\subsubsection{Individual vs Joint Learning of Signal Models} \label{SvsJ}

We lastly verify our theoretical results in Section \ref{sec_theo_anlys} by examining the performance of our method in the settings of individual model learning on single graphs and joint model learning on multiple graphs.  
We synthetically generate signals on two graphs $\G^1$ and $\G^2$ with  $N^1 = N^2 = 100$  nodes and $J=4$ kernels. The signals on graph $\G^1$ are formed with respect to the reference kernel parameter vector $\psivect_1$, whereas the signals on $\G^2$ are formed with respect to the parameter vector $\psivect_2$, which is a perturbed version of $\psivect_1$. The deviation between the two parameter vectors is set by tuning the spectrum discrepancy parameter $\delpsi=\| \psivect_2-\psivect_1 \|$. The missing observation ratio is fixed to $40\%$. The data size (number of signals) $K=K^1=K^2$ is varied within the range $[5, 200]$ and the NMSE of the missing observations on the reference graph is examined in the individual and joint learning settings.

\begin{figure}[t]
     \centering
     \begin{subfigure}[b]{0.23\textwidth}
         \centering
         \includegraphics[height=3.7cm]{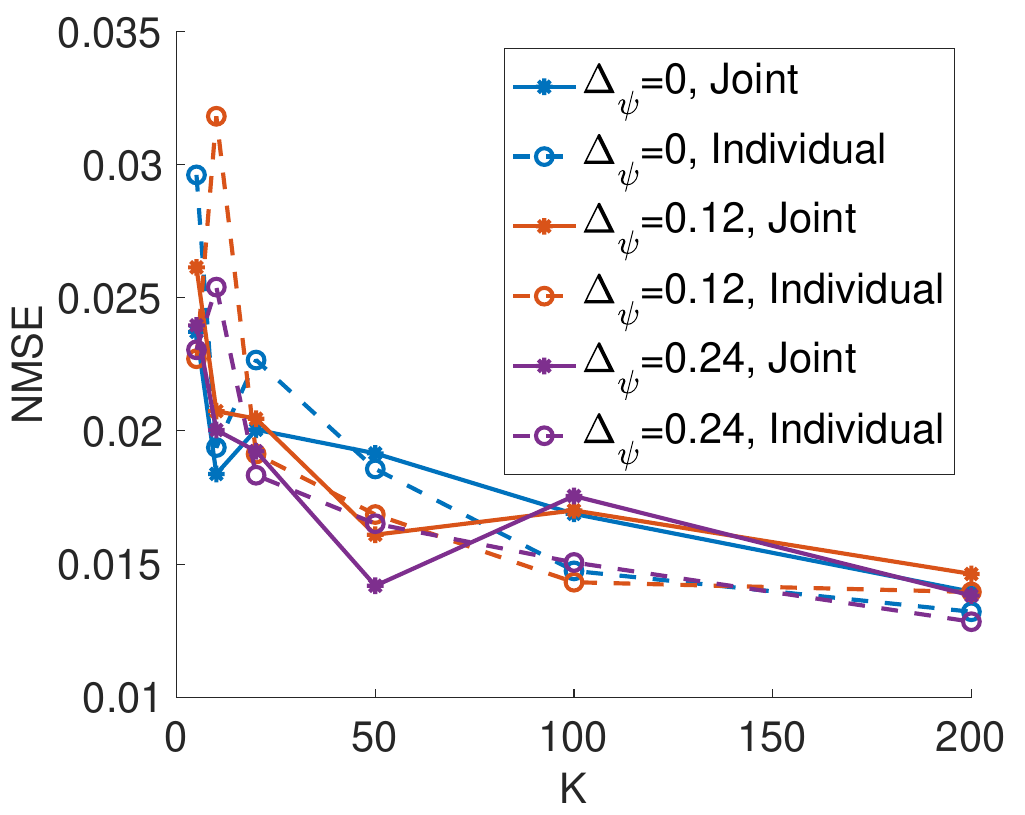}
         \caption{}
          \label{fig_NMSE_vs_K}
     \end{subfigure}
     \hfill
     \begin{subfigure}[b]{0.23\textwidth}
         \centering
         \includegraphics[height=3.7cm]{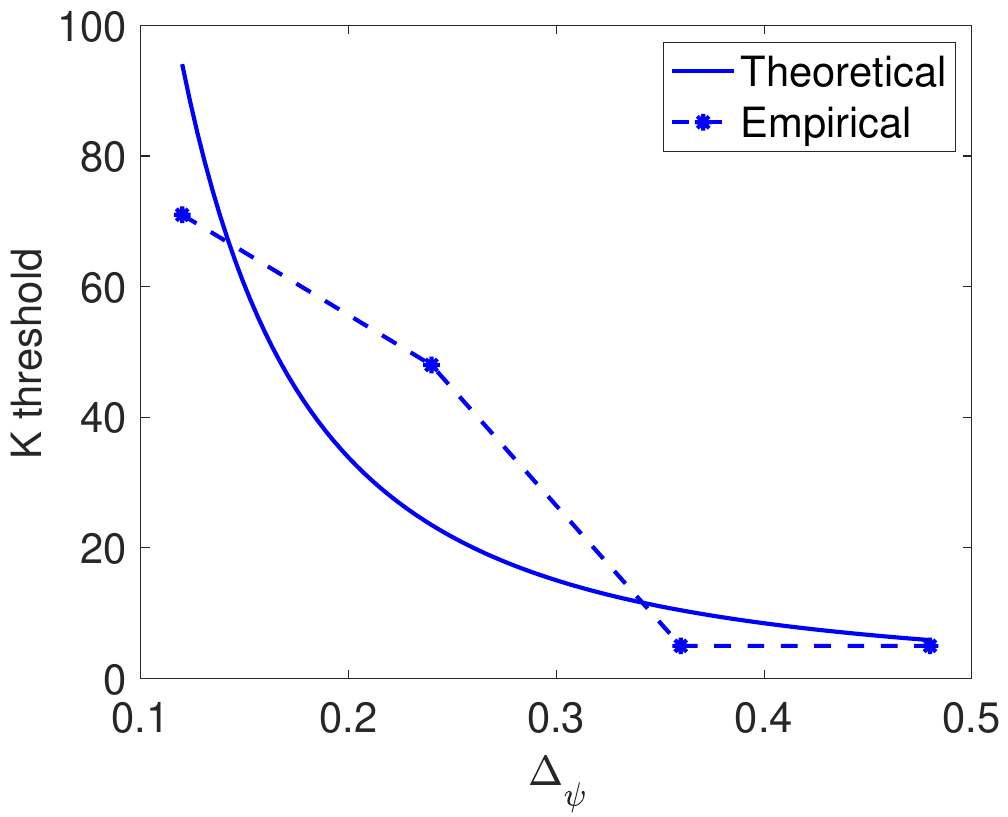}
         \caption{}
         \label{fig_K_vs_DeltaPsi}
     \end{subfigure}
     \hfill
	\caption{(a) Variation of the NMSE with the number of signals $K$. (b) The threshold value of $K$ under which joint learning outperforms individual learning.}
 \label{fig_eval_theo_bnds}
\end{figure}

In Figure \ref{fig_NMSE_vs_K}, we present the variation of the NMSE with the number of signals for various values of the spectrum discrepancy $\delpsi$. The estimation error has the general tendency to decrease with the data size $K$, in line with the result in Theorem \ref{thm_dev_unobs_obs}, which predicts the accuracy of estimation to improve at rate $1-O(1/K)$ as $K$ increases. While at small $K$ values joint learning is more advantageous than individual learning, as $K$ increases, individual learning is observed to outperform joint learning. We recall from Remark 1 that the critical $K$ value under which joint learning is favorable against individual learning decreases at rate $K=O(1/\delpsisq)$ as the spectral discrepancy increases. In order to study this behavior, in Figure \ref{fig_K_vs_DeltaPsi}, we examine how the threshold value of $K$ under which joint learning outperforms individual learning varies with the spectral discrepancy $\delpsi$ (the threshold $K$ value is found from smoothed versions of the curves in Figure \ref{fig_NMSE_vs_K} for numerical consistency).  The empirically obtained threshold for $K$ is clearly seen to decrease as $\delpsi $ increases, where we also present the theoretical prediction decaying at rate $K=O(1/\delpsisq)$ for visual comparison. These findings are aligned with the tradeoff pointed to by Corollary \ref{cor_exp_test_error}, stating that the negative impact of lack of data can be successfully compensated for by jointly learning a common model between different graph data sets, provided that the deviation between the spectral contents of data sets remains tolerable.


\subsection{Comparative Experiments} \label{compExp}

We next conduct comparative experiments where each graph signal $\yim$ has missing observations at a randomly selected subset of nodes, and the signal estimation performance of the proposed method (SGKL) is compared to that of the following baseline methods: The preliminary version of our method which solves the MAP estimation problem \eqref{eq_obj_final_map} without the additional regularization terms (SCGDL) \cite{TurhanV21};  spectral graph wavelet dictionaries\footnote{We reconstruct graph signals over wavelet dictionaries with the tight frame, Meyer, Mexican hat, and ab-spline kernels and report the smallest errors.}   (SGWT)  \cite{hammond2011wavelets}; graph-based signal interpolation via Tikhonov regularization (Gr-Tikhonov)  adapted from \cite{zhou2004learning}; graph-based total variation minimization (TVMin) \cite{JungHMJHE19}; non-smooth graph signal interpolation via  linear structural equation models (LSEM) \cite{MAZARGUIL2022108480};  optimal Papoulis-Gerchberg iterative reconstruction \cite{8918094} (O-PGIR); graph-enhanced multi-scale dictionary learning\footnote{As GEMS is not a standalone graph signal interpolation method, we couple it with the Gr-Tikhonov \cite{zhou2004learning}, TVMin \cite{JungHMJHE19}, and LSEM \cite{MAZARGUIL2022108480} methods for preprocessing the data prior to dictionary learning and report the results with the smallest errors.} (GEMS)  \cite{8642839}; and graph attention networks with dynamic attention coefficients (GATv2) \cite{Brody0Y22}. We set the number of kernels to the default value  $J=4$ used in \cite{hammond2011wavelets} for the SGWT, SCGDL, and SGKL methods.

\begin{figure}[t]
     \centering
     \begin{subfigure}[b]{0.22\textwidth}
         \centering
         \includegraphics[height=3.3cm]{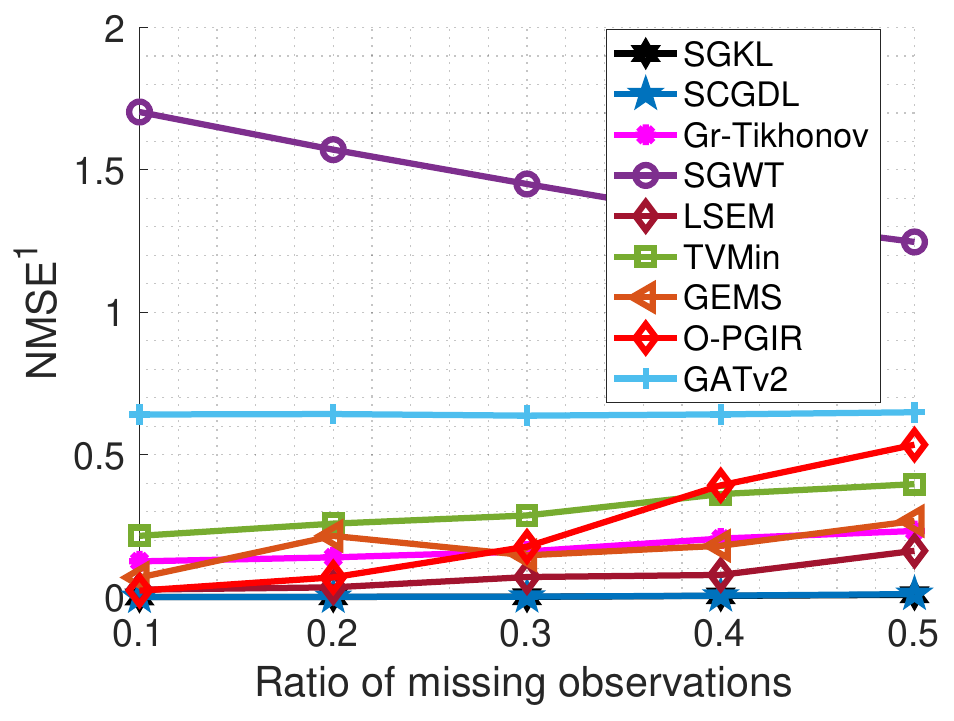}
         \caption{}
          \label{sg1}
     \end{subfigure}
     \hfill
     \begin{subfigure}[b]{0.22\textwidth}
         \centering
         \includegraphics[height=3.3cm]{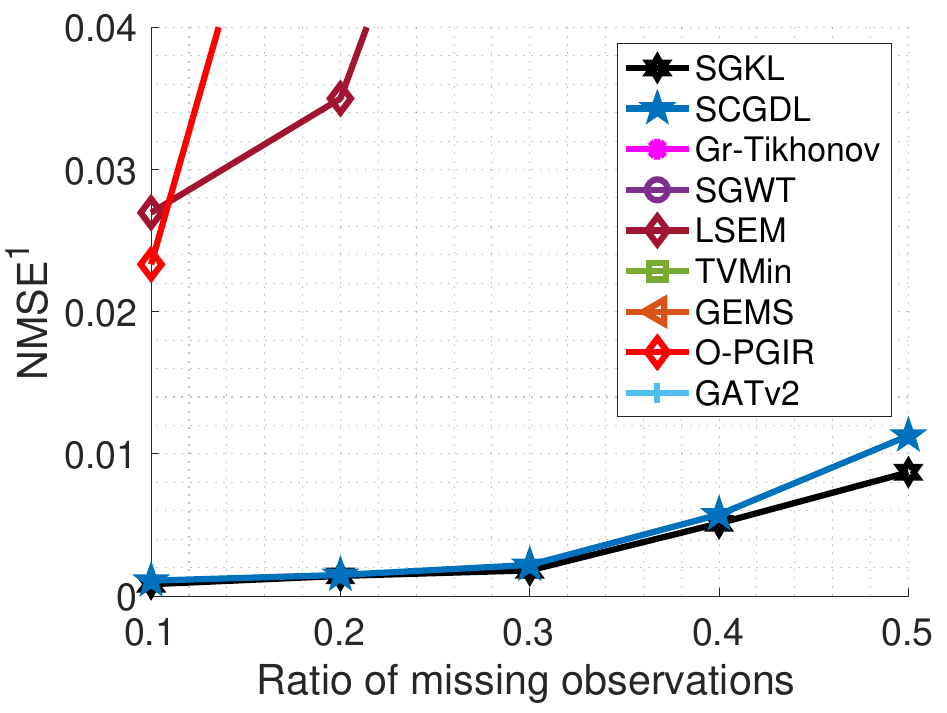}
         \caption{}
         \label{sg1z}
     \end{subfigure}
     \hfill
      \begin{subfigure}[b]{0.22\textwidth}
         \centering
         \includegraphics[height=3.3cm]{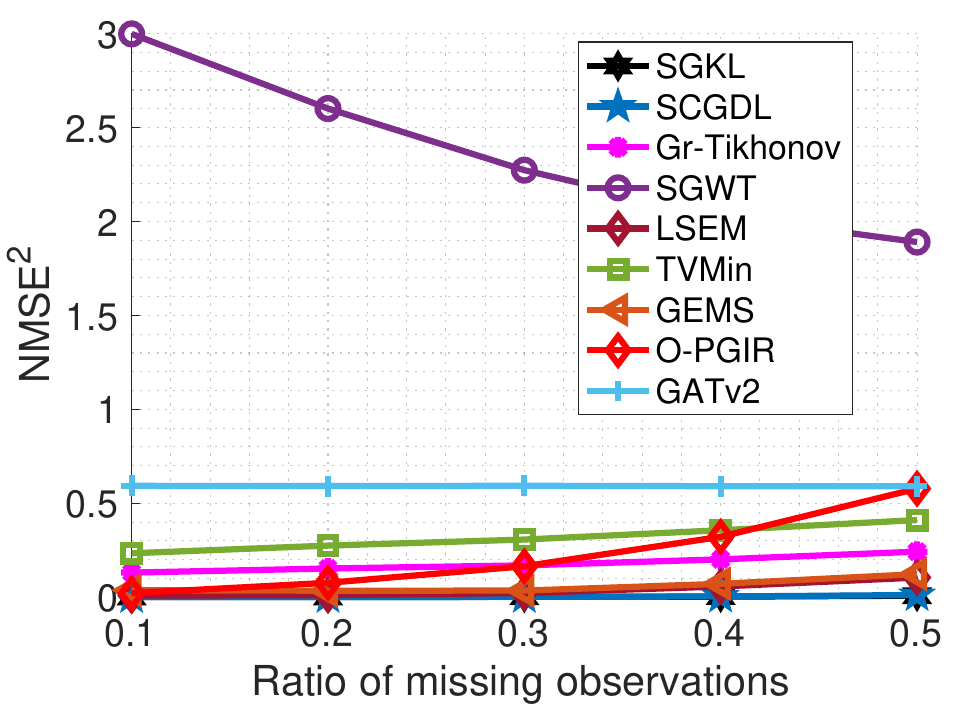}
         \caption{}
          \label{sg2}
     \end{subfigure}
     \hfill
 \begin{subfigure}[b]{0.22\textwidth}
         \centering
         \includegraphics[height=3.3cm]{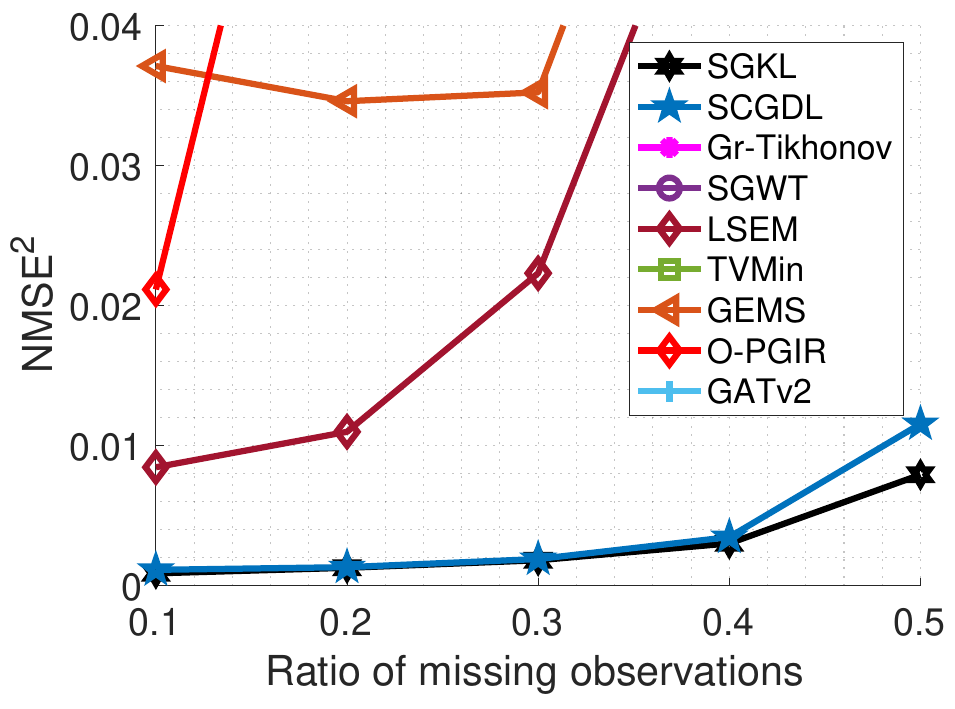}
         \caption{}
          \label{sg2z}
     \end{subfigure}
     \hfill
	\caption{Estimation errors of compared methods on (a), (b): $\mathcal{G}^1$ and  (c), (d): $\mathcal{G}^2$ for the synthetic data set. The results of the algorithms with the lowest errors are replotted in right panels (b), (d) for visual clarity.}
 \label{fig_NMSE_comp_syn}
\end{figure}

\begin{figure}[t]
     \centering
     \begin{subfigure}[b]{0.23\textwidth}
         \centering
         \includegraphics[height=3.3cm]{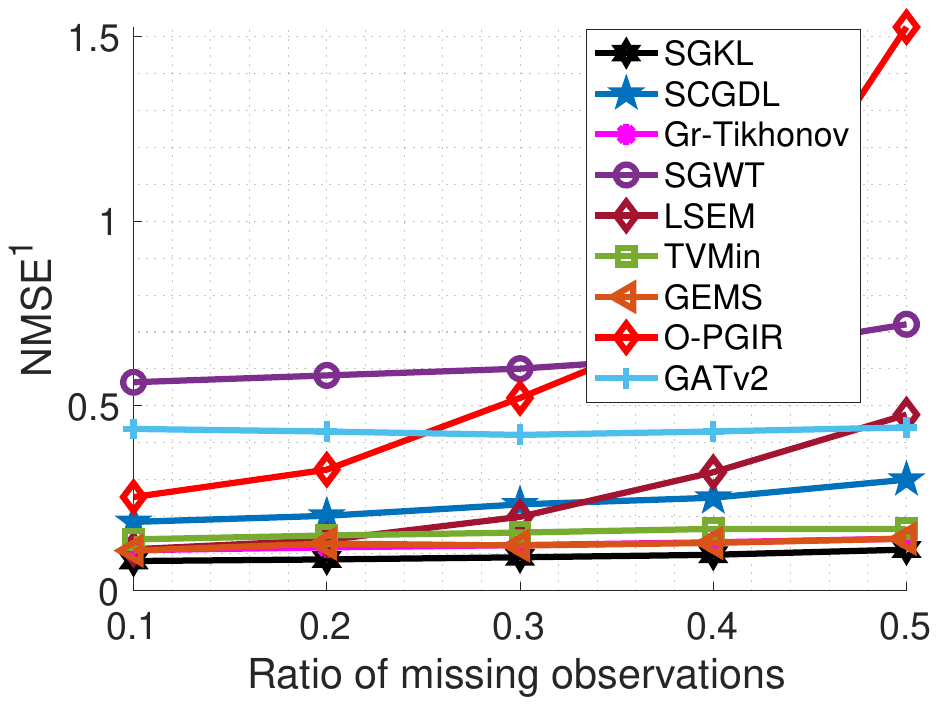}
         \caption{}
          \label{fig_molene_temp}
     \end{subfigure}
     \hfill
     \begin{subfigure}[b]{0.23\textwidth}
         \centering
         \includegraphics[height=3.3cm]{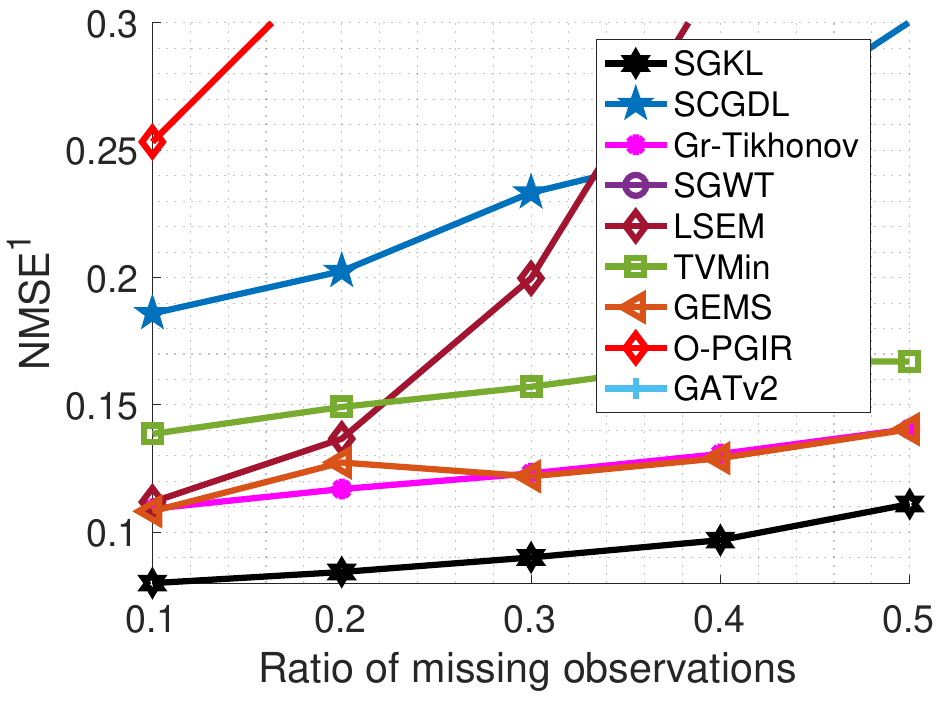}
         \caption{}
         \label{fig_molene_tempz}
     \end{subfigure}
     \hfill
      \begin{subfigure}[b]{0.23\textwidth}
         \centering
         \includegraphics[height=3.3cm]{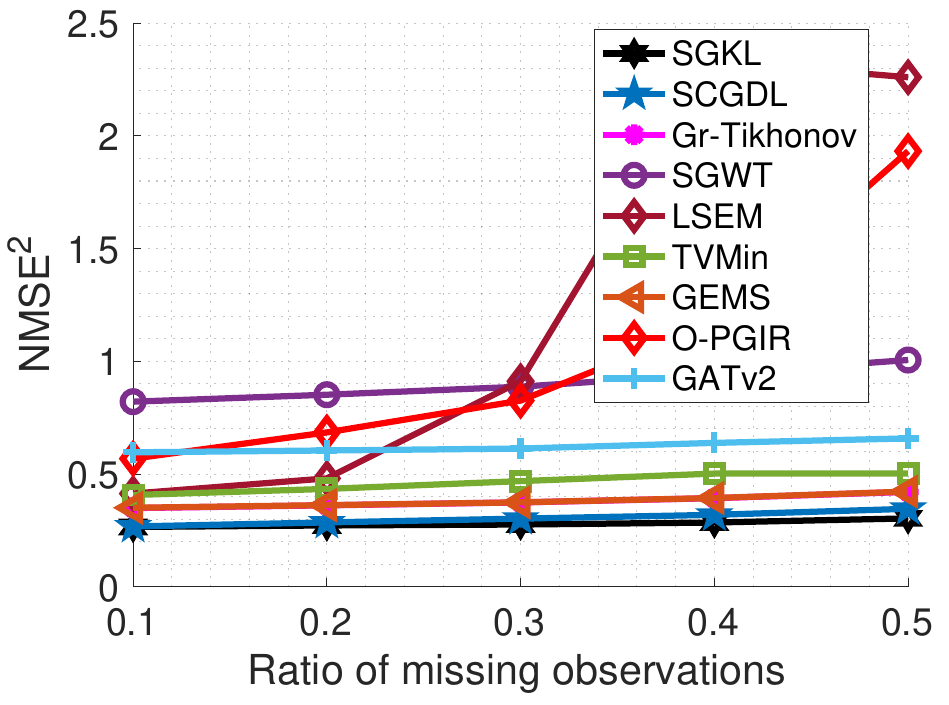}
         \caption{}
          \label{fig_molene_wind}
     \end{subfigure}
     \hfill
 \begin{subfigure}[b]{0.23\textwidth}
         \centering
         \includegraphics[height=3.3cm]{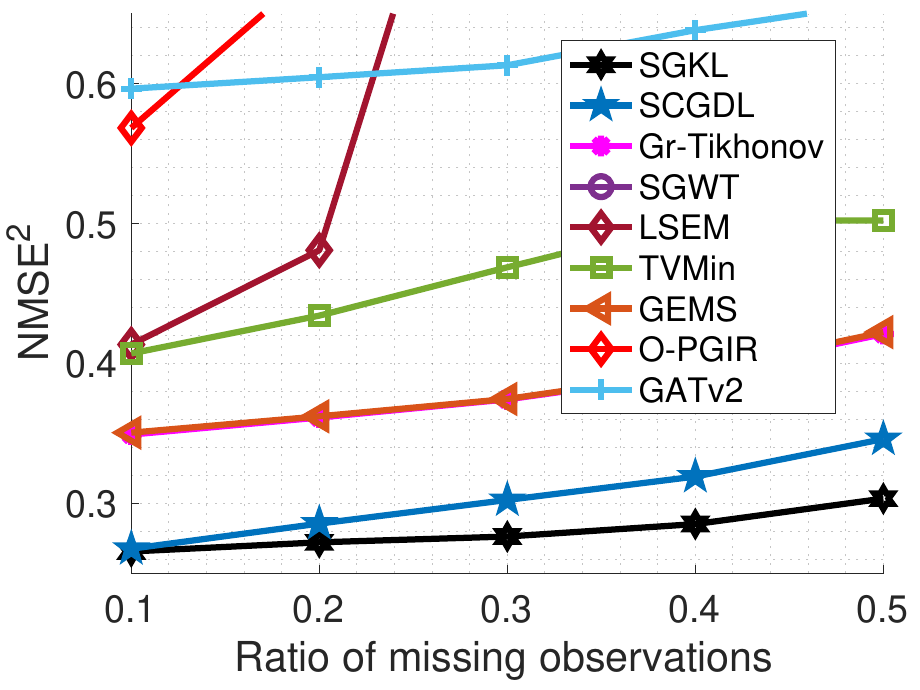}
         \caption{}
          \label{fig_molene_windz}
     \end{subfigure}
     \hfill
	\caption{Estimation errors of compared methods on (a), (b): $\mathcal{G}^1$ and  (c), (d): $\mathcal{G}^2$ for the Mol\`ene data set. The results of the algorithms with the lowest errors are replotted in right panels (b), (d) for visual clarity.}
 \label{fig_NMSE_comp_molene}
\end{figure}

\begin{figure}[t]
     \centering
     \begin{subfigure}[b]{0.23\textwidth}
         \centering
         \includegraphics[height=3.3cm]{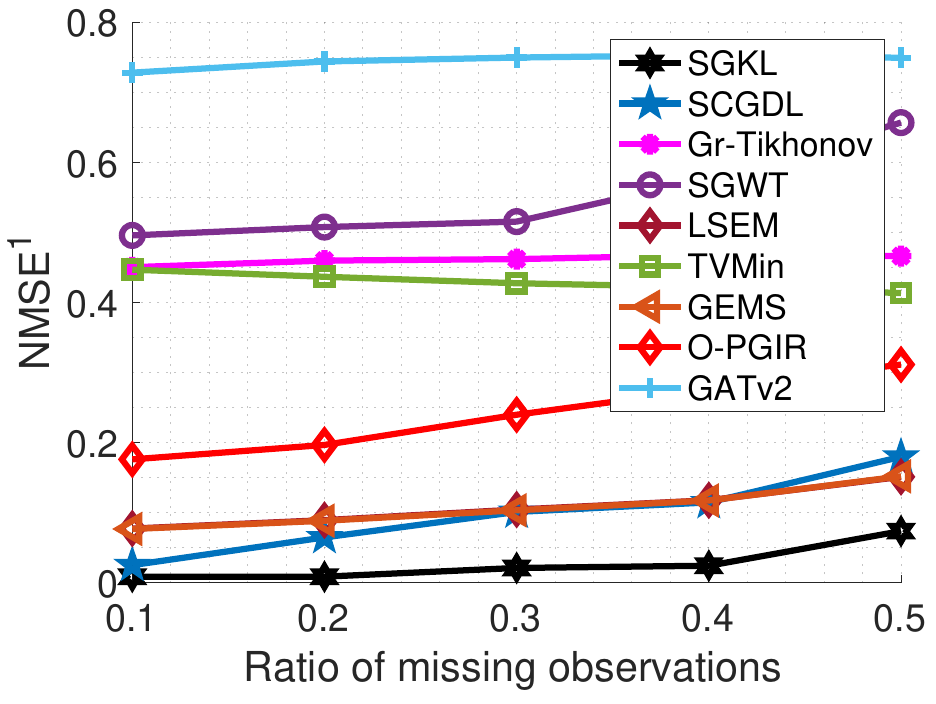}
         \caption{}
          \label{fig_covid_eu}
     \end{subfigure}
     \hfill
      \begin{subfigure}[b]{0.23\textwidth}
         \centering
         \includegraphics[height=3.3cm]{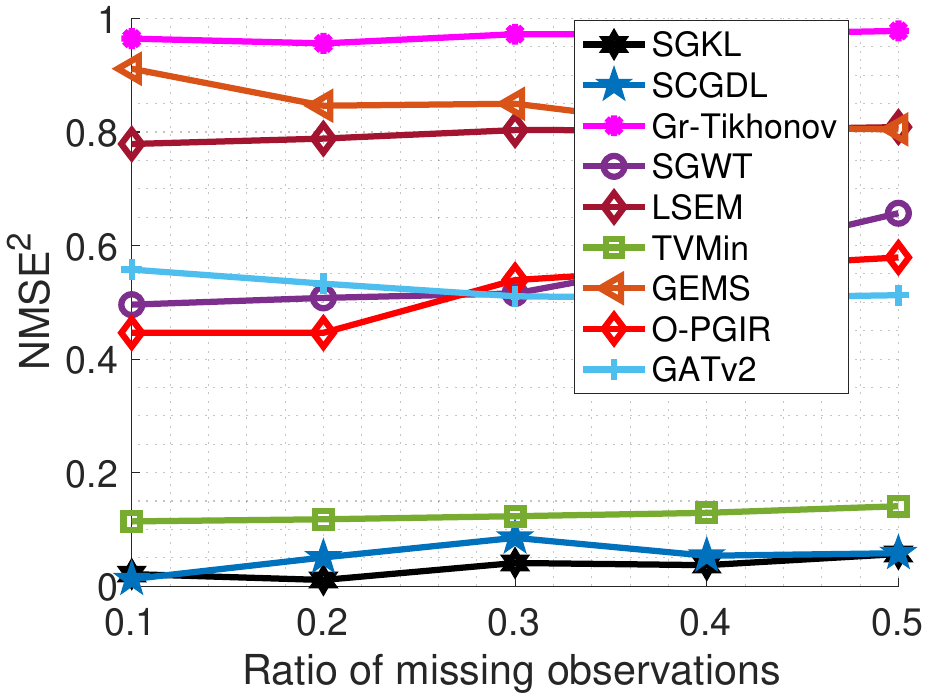}
         \caption{}
          \label{fig_covid_us}
     \end{subfigure}
     \hfill
	\caption{Estimation errors of compared methods on (a): $\mathcal{G}^1$ and  (b): $\mathcal{G}^2$ for the COVID-19 data set.}
 \label{fig_NMSE_comp_covid}
\end{figure}

\begin{figure}[t]
     \centering
     \begin{subfigure}[b]{0.23\textwidth}
         \centering
         \includegraphics[height=3.3cm]{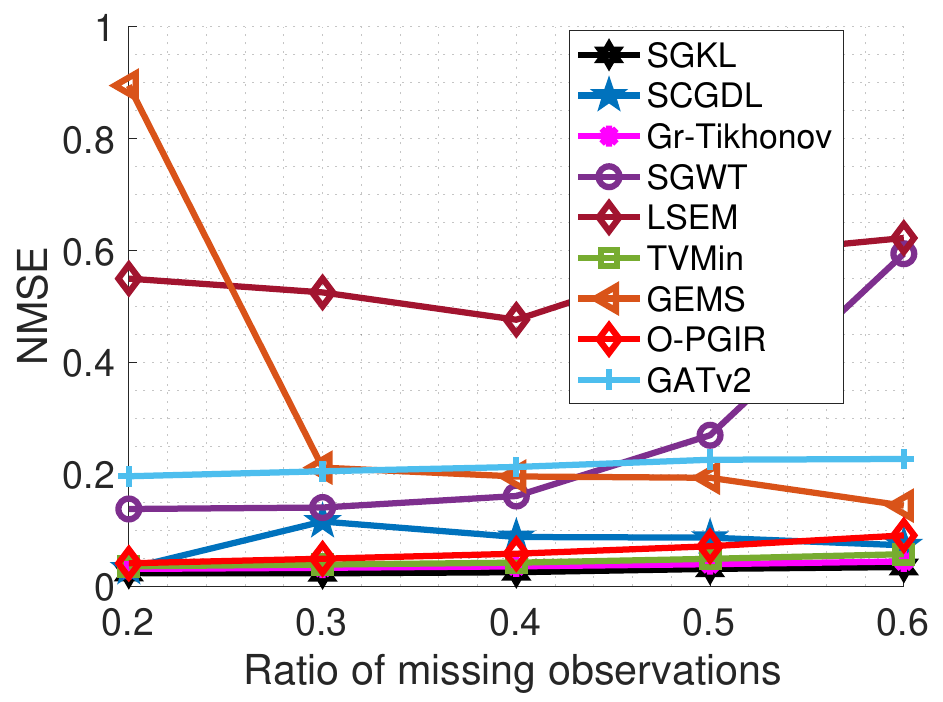}
         \caption{}
          \label{fig_noaa}
     \end{subfigure}
     \hfill
      \begin{subfigure}[b]{0.23\textwidth}
         \centering
         \includegraphics[height=3.3cm]{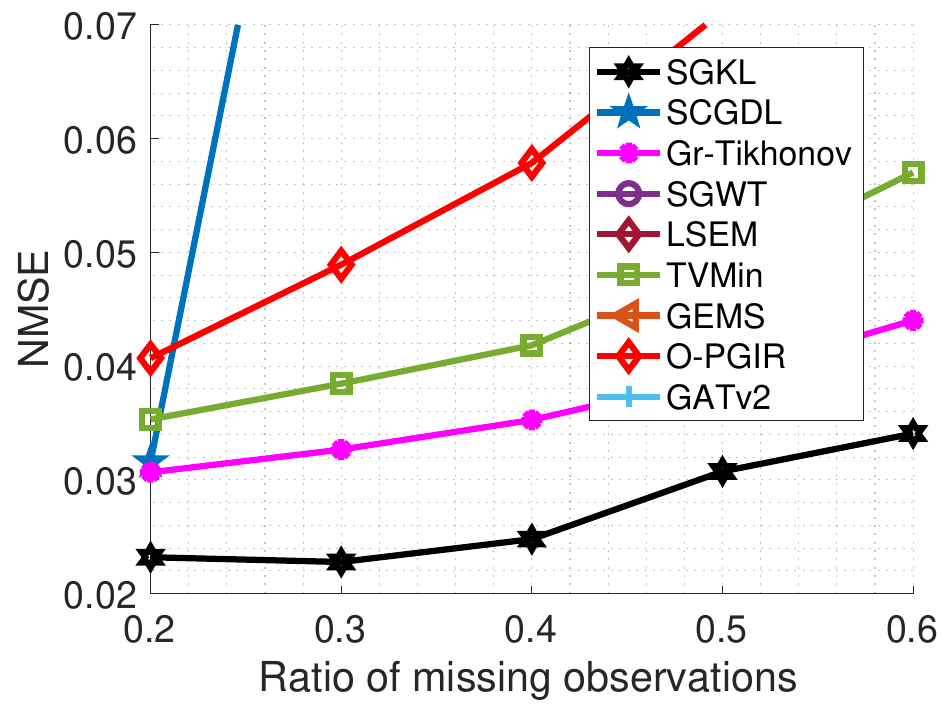}
         \caption{}
          \label{fig_noaaz}
     \end{subfigure}
     \hfill
	\caption{Estimation errors of compared methods on the NOAA data set (a). The results of the algorithms with the lowest errors are replotted in panel (b) for visual clarity.}
 \label{fig_NMSE_comp_noaa}
\end{figure}

The NMSE values of the compared methods are presented in Figures \ref{fig_NMSE_comp_syn}-\ref{fig_NMSE_comp_noaa}  for the synthetic, Mol\`ene, COVID-19, and the NOAA data sets, respectively. A general overview of the results shows that the proposed SGKL algorithm very often outperforms the other methods in terms of estimation error. For the synthetic data set in Figure \ref{fig_NMSE_comp_syn}, SGKL attains NMSE values very close to 0, confirming its efficiency in minimizing the objective function and learning accurate data models.  The minimal increase in its estimation error with increasing missing observation ratios suggests its robustness to the lack of data. The considerable performance gap between SCGDL and SGKL in Figures \ref{fig_NMSE_comp_molene}, \ref{fig_NMSE_comp_covid}, and \ref{fig_NMSE_comp_noaa} confirms that the triple regularization scheme employed in the proposed SGKL method improves the model estimation accuracy significantly. 

The comparison of the graph dictionary approaches among themselves  suggests that the SGKL and SCGDL methods, which learn from data, provide significantly better accuracy compared to SGWT, which solely uses the information of the graph topology and not the signal structure.  The GEMS algorithm achieves reconstruction error rates comparable to the corresponding graph-based interpolation algorithms it is initialized with. Relying on smoothness and band-limitedness assumptions, the Gr-Tikhonov and O-PGIR methods are seen to yield higher error than the proposed SGKL, overlooking the high-frequency spectral components of data. Learning complex nonlinear models from data, GATv2 outperforms the LSEM and O-PGIR methods in some settings, which are based on simpler signal models. The local reconstruction approach employed in the TVMin and LSEM methods may impair the reconstruction accuracy due to the dependency of the performance on the information of the neighboring nodes. Overall, the experiments demonstrate the ability of the proposed SGKL method to learn spectral  kernels that are successfully tuned to the common frequency characteristics of the whole signal collection, achieving quite accurate signal reconstruction results.

\section{Conclusion}
\label{sec_conclusion}

We have proposed a representation learning algorithm for graph signals based on the assumption that the signal energy is concentrated at only a few components of the graph spectrum, each of which is modeled with a spectral kernel. The learning problem is then formulated as the joint optimization of the kernel spectrum parameters along with the representation coefficients of the signals. The proposed algorithm allows the incorporation of data from multiple independently constructed graphs in the learning, whose influence on the reconstruction accuracy is studied with a detailed theoretical analysis. Experiments on several synthetic and real data sets demonstrate the reconstruction performance of the proposed method in comparison with various approaches in the literature, pointing to the potential of spectrally concentrated representations in the modeling of graph signals. The extension of our work to time-varying graph signals, dynamically changing graph topologies, and decentralized or online representation learning schemes are some of the potential future directions of our study.

\appendices
\section*{Appendix A: Computation of the gradient in the optimization of kernel parameters}

Here we derive the gradient expression used in the optimization of the kernel parameters $\psivect$ when solving \eqref{eq_init_obj13}. 

In order to find the gradient  $\frac{\partial f({\psivect})}{\partial {\psivect} }$ of the function $f(\psivect)$ in \eqref{eq_init_obj13}, we first find the gradient of the first three terms by following similar steps as proposed in \cite{TurhanV21}: The first two terms in $f({\psivect}) $  can be written as
\begin{eqnarray}
	f_{1,2}({\psivect})& = & (\psivect-\xivect)^T \Nmat (\psivect-\xivect) \nonumber .
\end{eqnarray}
Here, $ \xivect \in \mathbb{R}^{2J \times 1}$ is a vector containing the value $0$ and the predetermined nominal scale value $s_0$, respectively in its  first $J$ and last $ J $ elements. $ \Nmat \in \mathbb{R}^{2J \times 2J}  $ is a diagonal matrix whose first $ J $ diagonal elements are $ 1 $, and last $ J $ diagonal elements are $\eta_{s}$.
The gradient of $f_{1,2}({\psivect}) $ with respect to $ {\psivect} $ is
$
	\frac{\partial f_{1,2}({\psivect})}{\partial {\psivect} }=2 \Nmat \psivect- 2\Nmat \xivect.
$
In order to derive the gradient of the third term in \eqref{eq_init_obj13}, first we define
 $f_3(\psivect)= \sum_{m=1}^M \sum_{i=1}^{K^m} f_3^{m,i}(\psivect)$, where 
\begin{eqnarray*}
	f_3^{m,i}(\psivect)=\| \Sim \yim - \Sim \Dmpsi \xim \|^2 \nonumber \\ 
	=\| \Sim \yim - \Sim \Um \Gmat^m(\psivect)  (\Vmat^m)^T \xim \|^2 \nonumber.\\
\end{eqnarray*}
Here the matrix $ \Gmat^m(\psivect)=[
	\hat{g}_1(\Lammm)  \ \hat{g}_2(\Lamm^m) \  \cdots  \ \hat{g}_J(\Lammm) 
] \in \mathbb{R}^{N^m \times J N^m}$ consists of the filter kernels and the block diagonal matrix
$\Vmat^m \in \mathbb{R}^{J N^m \times J N^m}$ contains the matrix $\Um \in \mathbb{R}^{N^m \times N^m}$ on its diagonals $J$ times.
%

Defining the vector $\avect= \Sim \yim \in \mathbb{R}^{R_i^{m} \times 1}$, the matrix $\Cmat= \Sim \Um \in \mathbb{R}^{R_i^{m} \times N^m}$ and also $\cvect= (\Vmat^m)^T \xim \in \mathbb{R}^{J N^m \times 1}$, we can write 
$
	f_3^{m,i}(\psivect)=\| \avect- \Cmat \Gmat^m(\psivect) \cvect\|^2		
$, 
from which we get 
\begin{eqnarray}
	\frac{\partial f_3^{m,i}(\psivect)}{\partial {\Gmat^m(\psivect)} }&=2 \Cmat^T \Cmat \, \Gmat^m(\psivect) \, \cvect \cvect^T \nonumber
	-2 \Cmat^T  \avect \cvect^T  \in  \mathbb{R}^{N^m \times J N^m}   .\nonumber
\end{eqnarray}
Since the matrix $ \Gmat^m(\psivect) $ has only $ J N^m $ nonzero entries located on the diagonals of the diagonal matrices $ \hat{g}_1(\Lammm)$, $\hat{g}_2(\Lammm)$, $\cdots $,   $\hat{g}_J(\Lammm) $, we form a vector from $ \frac{\partial f_3^{m,i}(\psivect)}{\partial {\Gmat^m(\psivect)}} $ by extracting only its entries corresponding to these $ J N^m $ nonzero entries of $ \Gmat^m(\psivect) $, which we denote as $ \left( \frac{\partial f_3^{m,i}(\psivect)}{\partial {\Gmat^m(\psivect)} } \right)_v  \in \mathbb{R}^{JN^m\times1}$. We can then obtain the derivative of $f_3^{m,i}(\psivect) $ with respect to $ \psivect $ by applying the chain rule as 
\begin{equation}
	\Big(\frac{\partial f_3^{m,i}(\psivect)}{\partial {\psivect} }\Big)^T= \left( \frac{\partial f_3^{m,i}(\psivect)}{\partial {\Gmat^m(\psivect)} } \right)_v^T \frac{\partial {\Gmat^m(\psivect)}}{\partial \psivect }.
\label{eq:grad_f3m}
\end{equation}

We thus need to compute the derivative $ \frac{\partial {\Gmat^m(\psivect)}}{\partial \psivect }$. We first compute the partial derivatives of each subdictionary kernel $ \hat{g}_j(\lambda) $, for $ j=1,2, \dots, J $, with respect to all $ \mu_j $'s and $s_j$'s. We then form vectors $ \frac{\partial{\hat{g}_j^m}}{\partial{\mu_j}} $ and $ \frac{\partial{\hat{g}_j^m}}{\partial{s_j}} $ $ \in \mathbb{R}^{N^{m}\times1} $ by putting together the partial derivatives of the kernel functions $ \hat{g}_j(\lambda) $ evaluated at the eigenvalues of each graph Laplacian $\Lapm$. Thus arranging the partial derivatives of the kernel functions as below, we obtain the gradient  $ \frac{\partial {\Gmat^m(\psivect)}}{\partial \psivect } \in \mathbb{R}^{J N^m\times 2J}$ as follows:
\begin{equation*}
	\frac{\partial {\Gmat^m(\psivect)}}{\partial \psivect } = \begin{bmatrix}
			\frac{\partial{\hat{g}_1^m}}{\partial{\mu_1}} & 0 & \cdots & 0 & \frac{\partial{\hat{g}_1^m}}{\partial{s_1}} & 0 & \cdots & 0 \\ 
			0 & \frac{\partial{\hat{g}_2^m}}{\partial{\mu_2}} & \cdots & 0 & 0 & \frac{\partial{\hat{g}_2^m}}{\partial{s_2}} & \cdots & 0 \\
			0 & 0 & \ddots & 0 & 0 & 0 & \ddots & 0 \\
			\vdots & \vdots & \vdots & \vdots & \vdots & \vdots & \vdots & \vdots\\
			0 & 0 & \cdots & \frac{\partial{\hat{g}_J^m}}{\partial{\mu_J}} & 0 & 0 & \cdots & \frac{\partial{\hat{g}_J^m}}{\partial{s_J}} 
	\end{bmatrix}
\end{equation*}

Lastly, in order to find the gradient of the fourth term, let us define
\begin{equation}	
		f_4(\psivect)=\sum_{m=1}^{M} \tr((\Xm)^{T}(\Dmpsi)^{T} \Lapm \Dmpsi \Xm).
\label{eq:init_obj14}
\end{equation}
Again decomposing the dictionaries as $\Dmpsi =\Um \Gmat^m(\psivect)  (\Vmat^m)^T$, defining $\Fmat=(\Xm)^{T} \Vmat^m$, and recalling that  $(\Um)^{T} \Lapm \Um=\Lammm$, we get the derivative of $f_4(\psivect)$ with respect to $\Gmat^m(\psivect)$ as
\begin{equation}	
		\frac{\partial f_4({\psivect})}{\partial {\Gmat^m(\psivect)} }=2 \Lammm \Gmat^m(\psivect)\Fmat^T \Fmat. 
\label{eq:init_obj15}
\end{equation}

Forming a vector from $\partial f_4({\psivect})/ \partial {\Gmat^m(\psivect)}$ by arranging its entries corresponding to the nonzero entries of  $\Gmat^m(\psivect)$, and denoting it as 
 $\left( \frac{\partial f_4({\psivect})}{\partial {\Gmat^m(\psivect)} } \right)_v \in \mathbb{R}^{J N^m \times 1}$, we get the gradient of $f_4(\psivect)$ as
\begin{equation}	
	\begin{split}
		&\left(\frac{\partial f_4({\psivect})}{\partial {\psivect} }\right)^{T}
		=\left(\frac{\partial f_4({\psivect})}{\partial {\Gmat^m(\psivect)} }\right)_v^{T}\frac{\partial \Gmat^m({\psivect})}{\partial {\psivect} }.
	\end{split}
\label{eq:init_obj16}
\end{equation}

Finally, putting together all these terms, we get
\begin{eqnarray*}
	\frac{\partial f(\psivect)}{\partial {\psivect} }&=&\frac{\partial f_{1,2}({\psivect})}{\partial {\psivect} }+\eta_{w} \sum_{m=1}^{M} \sum_{i=1}^{K^m} \frac{\partial f_3^{m,i}(\psivect)}{\partial {\psivect} }
	+ \eta_{y} \frac{\partial f_4({\psivect})}{\partial {\psivect} }.
\end{eqnarray*}

\section*{Appendix B: Proof of Theorem \ref{thm_dev_unobs_obs}}

We first present the following Lemma, which will be useful in the proof of Theorem \ref{thm_dev_unobs_obs}.

\begin{lemma}
\label{lem_var_avgerrtrain}
Let $\avgerrtrain(\psivect_0) \triangleq  \frac{1}{M} \sum_{m=1}^M \frac{1}{\Km} \sum_{i=1}^\Km \errtrainim(\psivect_0)$ denote the average approximation error of the observed signal components when represented in the dictionary $D(\psivect_0)$ for a given $\psivect_0 \in \Psispace $. Then, for any $\psivect_0 \in \Psispace $, the variance of $\avgerrtrain(\psivect_0)$ is bounded as
\begin{equation*}
\begin{split}
\var(\avgerrtrain(\psivect_0) ) \leq  \frac{1}{M^2} \sum_{m=1}^M  \frac{\cparm}{\Km}.
\end{split}
\end{equation*}
\end{lemma}
The proof of Lemma \ref{lem_var_avgerrtrain} is given in Appendix D. We now proceed with the proof of Theorem \ref{thm_dev_unobs_obs}:
\begin{proof}

We first observe that the expected average estimation error of the unobserved samples can be bounded as
\begin{equation}
\label{eq_bnd_testim_trainm_thm1}
\begin{split}
&E \left[ \frac{1}{M} \sum_{m=1}^M \frac{1}{\Km} \sum_{i=1}^\Km \errtestim(\psiest) \right] 
= \frac{1}{M} \sum_{m=1}^M \frac{1}{\Km} \sum_{i=1}^\Km E[\errtestim(\psiest) ] \\
&\leq \frac{1}{M} \sum_{m=1}^M E[\errtestm(\psiest)] 
\leq \frac{1}{M} \sum_{m=1}^M E[\errtrainm(\psiest)] + \csmooth
\end{split}
\end{equation}
where the first and the second inequalities follow from \eqref{eq_assum_errdata_smaller} and \eqref{eq_assum_csmoothm} respectively. We then would like to bound the average value of $E[\errtrainm(\psiest)]$ in terms of the average value of the approximation error $\errtrainim(\psiest)$ of the observed components of the signals used in learning the model. Due to the compactness of the parameter domain $\Psispace$, there exists some $\psivect_k \in \covcents $ with $\dist(\psiest, \psivect_k) < \epsilon$. We have
\begin{equation}
\label{eq_main_ineq_testtraindev}
\begin{split}
& \left |  \frac{1}{M} \sum_{m=1}^M E[\errtrainm(\psiest)]  
-  \frac{1}{M} \sum_{m=1}^M \frac{1}{\Km} \sum_{i=1}^\Km \errtrainim(\psiest)   \right | \\
& \leq 
\left | \frac{1}{M} \sum_{m=1}^M E[\errtrainm(\psiest)]   -  \frac{1}{M} \sum_{m=1}^M E[\errtrainm(\psivect_k)]  \right |  \\
&+ 
\left | \frac{1}{M} \sum_{m=1}^M E[\errtrainm(\psivect_k)]  - 
\frac{1}{M} \sum_{m=1}^M \frac{1}{\Km} \sum_{i=1}^\Km  \errtrainim(\psivect_k) 
 \right | \\
 & +
 \left |   \frac{1}{M} \sum_{m=1}^M \frac{1}{\Km} \sum_{i=1}^\Km  \errtrainim(\psivect_k) -
  \frac{1}{M} \sum_{m=1}^M \frac{1}{\Km} \sum_{i=1}^\Km  \errtrainim(\psiest)
 \right |.
\end{split}
\end{equation}

In the sequel, we study each one of the three terms in the right hand side of \eqref{eq_main_ineq_testtraindev}. The first term in \eqref{eq_main_ineq_testtraindev} is bounded as 
\begin{equation}
\label{eq_bnd_term1_thm1}
\begin{split}
& \left | \frac{1}{M} \sum_{m=1}^M E[\errtrainm(\psiest)]   -  \frac{1}{M} \sum_{m=1}^M E[\errtrainm(\psivect_k)]  \right |  \\
& \leq \frac{1}{M} \sum_{m=1}^M  \big | \,  E[\errtrainm(\psiest) - \errtrainm(\psivect_k)]  \, \big | \\
& \leq \frac{1}{M} \sum_{m=1}^M    E[ \, | \errtrainm(\psiest) - \errtrainm(\psivect_k) | \, ]  
\leq \dist(\psiest, \psivect_k) < \epsilon
\end{split}
\end{equation}
where the second and third inequalities follow respectively from Jensen's inequality and the definition of the distance $\dist$ in \eqref{eq_defn_distance_d}. 

We next derive a probabilistic upper bound for the second term in \eqref{eq_main_ineq_testtraindev}. Recalling the definition in Lemma \ref{lem_var_avgerrtrain},  we notice that
\begin{equation*}
\begin{split}
\frac{1}{M} \sum_{m=1}^M \frac{1}{\Km} \sum_{i=1}^\Km \errtrainim(\psivect_k) = \avgerrtrain(\psivect_k) 
\end{split}
\end{equation*}
while the expectation of this expression corresponds to
\begin{equation*}
\begin{split}
E[ \avgerrtrain(\psivect_k) ] = \frac{1}{M} \sum_{m=1}^M E[\errtrainm(\psivect_k)] 
\end{split}
\end{equation*}
since $\errtrainim(\psivect_k)$ and $\errtrainm(\psivect_k)$ are i.i.d.~as justified in the proof of Lemma \ref{lem_var_avgerrtrain}. From Chebyshev's inequality and the upper bound on $\var(\avgerrtrain(\psivect_k) )$ presented in Lemma \ref{lem_var_avgerrtrain}, for any $\varepsilon>0$, for the particular parameter vector $\psivect_k $ we have
\begin{equation*}
\begin{split}
&P( | \avgerrtrain(\psivect_k)  - E[\avgerrtrain(\psivect_k) ]  | \geq \varepsilon) \leq \frac{\var(\avgerrtrain(\psivect_k) )}{\varepsilon^2}\\
& \leq \frac{1}{\varepsilon^2 M^2} \sum_{m=1}^M  \frac{\cparm}{\Km}.
\end{split}
\end{equation*}
Then from union bound it follows that with probability at least 
\begin{equation*}
\begin{split}
1- \frac{\covnumeps}{\varepsilon^2 M^2} \sum_{m=1}^M  \frac{\cparm}{\Km},
\end{split}
\end{equation*}
we have $ | \avgerrtrain(\psivect_l)  - E[\avgerrtrain(\psivect_l) ]  | < \varepsilon$ for all $\psivect_l \in \covcents $, so that for any possible solution $\psiest$ of the algorithm, the second term in the right hand side of \eqref{eq_main_ineq_testtraindev} is bounded as 
\begin{equation}
\label{eq_bnd_term2_thm1}
\begin{split}
\left | \frac{1}{M} \sum_{m=1}^M E[\errtrainm(\psivect_k)]  - 
\frac{1}{M} \sum_{m=1}^M \frac{1}{\Km} \sum_{i=1}^\Km  \errtrainim(\psivect_k) 
 \right | < \varepsilon.
\end{split}
\end{equation}
Finally, we study the third term in \eqref{eq_main_ineq_testtraindev}. We first observe that the expectation of this term can be bounded as 
%
%
\begin{equation*}
\begin{split}
& E[ \, | \avgerrtrain(\psivect_k) -  \avgerrtrain(\psiest)| \, ]  \\
&= E 
\left[
 \left |   \frac{1}{M} \sum_{m=1}^M \frac{1}{\Km} \sum_{i=1}^\Km  \errtrainim(\psivect_k) -
  \frac{1}{M} \sum_{m=1}^M \frac{1}{\Km} \sum_{i=1}^\Km  \errtrainim(\psiest)
 \right |
\right] \\
& \leq
   \frac{1}{M} \sum_{m=1}^M \frac{1}{\Km} \sum_{i=1}^\Km
   E[ |  \errtrainim(\psivect_k) - \errtrainim(\psiest) | ]
   \leq \dist(\psivect_k, \psiest) 
   < \epsilon.
\end{split}
\end{equation*}
Using this bound together with Markov's inequality, we have
\begin{equation}
\label{eq_bnd_term3_thm1}
\begin{split}
P(| \avgerrtrain(\psivect_k) -  \avgerrtrain(\psiest)| \geq \varepsilon ) 
\leq \frac{E[ \, | \avgerrtrain(\psivect_k) -  \avgerrtrain(\psiest)| \, ]} {\varepsilon}
< \frac{\epsilon}{\varepsilon}
\end{split}
\end{equation}
for the same $\varepsilon$ used in \eqref{eq_bnd_term2_thm1}. Now, choosing $\epsilon = \cprob \varepsilon $ and putting together the results in \eqref{eq_bnd_term1_thm1}, \eqref{eq_bnd_term2_thm1}, \eqref{eq_bnd_term3_thm1}, we conclude that with probability at least
\begin{equation*}
\begin{split}
1-  \cprob - \frac{\covnumcprobeps}{\varepsilon^2 M^2} \sum_{m=1}^M  \frac{\cparm}{\Km} 
\end{split}
\end{equation*}
the expression in \eqref{eq_main_ineq_testtraindev} is bounded as
\begin{equation*}
\begin{split}
& \left |  \frac{1}{M} \sum_{m=1}^M E[\errtrainm(\psiest)]  
-  \frac{1}{M} \sum_{m=1}^M \frac{1}{\Km} \sum_{i=1}^\Km \errtrainim(\psiest)   \right |
<  (2 + \cprob) \varepsilon.
\end{split}
\end{equation*}
Combining this with the inequality in \eqref{eq_bnd_testim_trainm_thm1}, we get the result stated in the theorem.
\end{proof}

\section*{Appendix C: Proof of Theorem \ref{thm_exp_emp_err}}

\begin{proof}
We first present the following lemmas, which will be useful in the proof of the theorem.

\begin{lemma}
\label{lem_Dm_Lipsch}
The deviation between the dictionaries generated by two parameter vectors $\psione, \psitwo \in \Psispace$ is upper bounded as   
 \begin{equation*}
\begin{split}
\| \Dm (\psitwo) - \Dm (\psione) \|_F^2 \leq  J \Nm \cDsq \| \psitwo - \psione \|^2.
\end{split}
\end{equation*}
for a constant $\cD>0$ and for all $m=1, \dots, M$.
\end{lemma}

\begin{lemma}
\label{lem_finite_mom_coeff}
There exists a finite constant $\cx>0$ such that for all $m=1, \dots, M$
\begin{equation*}
\sqrt{E[\| \ximest(\psiest) \|^2]} \leq \cx 
\ \text{ and } \
\sqrt{E[\| \xim - \ximest(\psiest) \|^2]} \leq \cx.
\end{equation*}
\end{lemma}

The proofs of Lemmas \ref{lem_Dm_Lipsch} and \ref{lem_finite_mom_coeff} are given in Appendices E and F. We are now ready to prove Theorem \ref{thm_exp_emp_err}. Let us define the functions $\fone, \ftwo : \Psispace \rightarrow \R$
\begin{equation*}
\begin{split}
\fone(\psivect) &\triangleq  \sum_{j=1}^{J} (\mu_j)^2+\eta_{s}\sum_{j=1}^{J}(s_j-s_0)^2 \\
\ftwo(\psivect) & \triangleq \eta_{w}\sum_{m=1}^{M}\sum_{i=1}^{\Km}\norm{ \Sim \yim- \Sim \Dmpsi \ximest }^2
\end{split}
\end{equation*}
where we set $\ximest = \ximest(\psiest) $. From the optimality of $\psiest$ due to \eqref{eq_defn_psiest}, we have
\begin{equation*}
\begin{split}
\fone(\psiest) + \ftwo(\psiest) \leq \fone(\psicom) + \ftwo(\psicom)
\end{split}
\end{equation*}
which gives
\begin{equation}
\label{eq_f2_bnd_pf_thm2}
\begin{split}
& \ftwo(\psiest) = \eta_{w}\sum_{m=1}^{M}\sum_{i=1}^{\Km}\norm{ \Sim \yim- \Sim \Dmpsiest \ximest }^2 \\
& \leq | \fone(\psicom) - \fone(\psiest) | \\
 &+ \eta_{w}\sum_{m=1}^{M}\sum_{i=1}^{\Km}\norm{ \Sim \yim- \Sim \Dmpsicom \ximest }^2 .
\end{split}
\end{equation}
We can then bound the total approximation error of observed signals as
\begin{equation}
\label{eq_ineq_totap_app_err}
\begin{split}
&\sum_{m=1}^{M}\sum_{i=1}^{\Km}\norm{ \Sim \yim- \Sim \Dmpsiest \ximest } \\
&\leq
\sqrt{\Ktot} \left( \sum_{m=1}^{M}\sum_{i=1}^{\Km}\norm{ \Sim \yim- \Sim \Dmpsiest \ximest }^2 \right)^{\frac{1}{2}} \\
&\leq
\sqrt{\Ktot} \bigg(
\frac{1}{\eta_w} | \fone(\psicom) - \fone(\psiest) |  \\
& \quad + \sum_{m=1}^{M}\sum_{i=1}^{\Km} \norm{ \Sim \yim- \Sim \Dmpsicom \ximest }^2
\bigg)^{\frac{1}{2}}  \\
& \leq 
\left(\frac{\Ktot}{\eta_w}\right)^{\frac{1}{2}}
 | \fone(\psicom) - \fone(\psiest) |^{\frac{1}{2}}   \\
& \quad +\sqrt{\Ktot}  \bigg( \sum_{m=1}^{M}\sum_{i=1}^{\Km} \norm{ \Sim \yim- \Sim \Dmpsicom \ximest }^2
\bigg)^{\frac{1}{2}} 
\end{split}
\end{equation}
where $\Ktot \triangleq \sum_{m=1}^M \Km$ denotes the total number of graph signals, the first inequality is due to the equivalence of $\ell_1$ and $\ell_2$ norms, and the second inequality follows from \eqref{eq_f2_bnd_pf_thm2}. 

We proceed by upper bounding the expectations of the two terms in the last inequality in \eqref{eq_ineq_totap_app_err}. For the first term, due to the compactness of the domain $\Psispace$ and the continuity of the function $\fone$, the difference $ | \fone(\psicom) -  \fone(\psiest) |$ is bounded; hence, there exists a constant $\cf$ such that 
\begin{equation}
\label{eq_cf_bnd_pfthm2}
\begin{split}
E\left[   | \fone(\psicom) -  \fone(\psiest) |^{1/2}  \right] \leq \cf.
\end{split}
\end{equation}
Next, for the second term, we have
\begin{equation}
\label{eq_pfthm2_jensens}
\begin{split}
&E\left[ 
 \bigg( \sum_{m=1}^{M}\sum_{i=1}^{\Km} \norm{ \Sim \yim- \Sim \Dmpsicom \ximest }^2
\bigg)^{\frac{1}{2}} 
 \right] \\
& \leq 
 \bigg(
 \sum_{m=1}^{M}\sum_{i=1}^{\Km} E \left[ \norm{ \Sim \yim- \Sim \Dmpsicom \ximest }^2 \right]
 \bigg)^{\frac{1}{2}} 
\end{split}
\end{equation}
due to Jensen's inequality and the concavity of the square root function. The term inside the expectation can be upper bounded as
\begin{equation}
\label{eq_three_terms_pfthm2}
\begin{split}
&\| \Sim \yim- \Sim \Dmpsicom \ximest \|^2 
= 
\| \Sim \yim 
- \Sim \Dmpsim \xim  \\
&+ \Sim \Dmpsim \xim 
 - \Sim \Dmpsicom \ximest \|^2 \\
 & \leq
2 \|  \Sim \yim 
- \Sim \Dmpsim \xim  \|^2 \\
 &+
 2 \|  \Sim \Dmpsim \xim 
 - \Sim \Dmpsicom \ximest  \|^2 \\
 &= 
 2 \|  \Sim \yim 
- \Sim \Dmpsim \xim  \|^2 \\
&+
 2 \|  \Sim \Dmpsim \xim 
 - \Sim \Dmpsim \ximest  \\
& + \Sim \Dmpsim \ximest 
 - \Sim \Dmpsicom \ximest  \|^2 \\
 &\leq
  2 \|  \Sim \yim 
- \Sim \Dmpsim \xim  \|^2 \\
&+
 4 \|  \Sim \Dmpsim \xim 
 - \Sim \Dmpsim \ximest \|^2 \\
& + 4 \| \Sim \Dmpsim \ximest 
 - \Sim \Dmpsicom \ximest  \|^2. \\
\end{split}
\end{equation}
We continue by studying the expectations of the rightmost three terms in \eqref{eq_three_terms_pfthm2}. The first term yields
\begin{equation}
\label{eq_first_term_pfthm2}
\begin{split}
&E[ \, 2 \|  \Sim \yim 
- \Sim \Dmpsim \xim  \|^2 \,] 
= E[ \, 2 \|  \Sim \wim   \|^2 \,] \\
&= 2 E \left [
\sum_{l=1}^\Rim \left( \wim(n_{i,l}^{m}) \right)^2
\right]
= 2  \sigma^2 E[\Rim] = 2  \sigma^2 \Theta(\Rc)
\end{split}
\end{equation}
where $\{n_{i,l}^{m}\}$ denote the indices of the observed entries of each signal $ y_i^m$; the third equality is due to the independence of $\Rim$ and $\wim$; and the fourth equality follows from the considered setting $\Rim = \Theta(\Rc)$.

Next, for the second term in \eqref{eq_three_terms_pfthm2} we have
\begin{equation*}
\begin{split}
&4 \, \|  \Sim \Dmpsim \xim 
 - \Sim \Dmpsim \ximest \|^2 \\
 &\leq 4 \,
 \| \Sim \Dmpsim \|_F^2 \, \| \xim - \ximest \|^2 
 \leq
 4 J \Nm \| \xim - \ximest \|^2 
\end{split}
\end{equation*}
where the bound on the norm of $\Dmpsim $ is obtained similarly to \eqref{eq_Dm_diff} in the last inequality. Taking expectation, we get 
%
%
\begin{equation}
\label{eq_second_term_pfthm2}
\begin{split}
&E[\, 4 \, \|  \Sim \Dmpsim \xim 
 - \Sim \Dmpsim \ximest \|^2 \,] \\
 &\leq 4 J \Nm
  E[  \| \xim - \ximest \|^2  ]
  \leq  4 J \Nm \cxsq
\end{split}
\end{equation}
where the last inequality is due to Lemma \ref{lem_finite_mom_coeff}. Similarly, for the third term in  \eqref{eq_three_terms_pfthm2}, we obtain
\begin{equation*}
\begin{split}
& 4 \| \Sim \Dmpsim \ximest 
 - \Sim \Dmpsicom \ximest  \|^2 \\
 &\leq 4 \| \Dmpsim - \Dmpsicom \|_F^2 \, \| \ximest \|^2 \\
 & \leq 4 J \Nm \cDsq \| \psim - \psicom \|^2 \| \ximest \|^2 
 \leq 4 J \Nm \cDsq   \delpsisq \| \ximest \|^2
\end{split}
\end{equation*}
where the second inequality is due to Lemma \ref{lem_Dm_Lipsch}.
Then, from Lemma \ref{lem_finite_mom_coeff}, the expectation of the third term in \eqref{eq_three_terms_pfthm2} can be bounded as
\begin{equation}
\label{eq_third_term_pfthm2}
\begin{split}
& E[\, 4 \| \Sim \Dmpsim \ximest 
 - \Sim \Dmpsicom \ximest  \|^2] \\
 &\leq 4 J \Nm \cDsq   \delpsisq E[ \| \ximest \|^2] 
 \leq 4 J \Nm \cDsq   \delpsisq \cxsq.
\end{split}
\end{equation}
Combining the bounds \eqref{eq_first_term_pfthm2}-\eqref{eq_third_term_pfthm2} with the results in \eqref{eq_pfthm2_jensens} and \eqref{eq_three_terms_pfthm2}, we obtain
\begin{equation*}
\begin{split}
&E\left[ 
 \bigg( \sum_{m=1}^{M}\sum_{i=1}^{\Km} \norm{ \Sim \yim- \Sim \Dmpsicom \ximest }^2
\bigg)^{\frac{1}{2}} 
 \right] \\
& \leq
 \bigg(
  \sum_{m=1}^{M}
2  \sigma^2 \Km \Theta(\Rc)
+
4 J \Nm \Km \cxsq \\
&+
4 J \Nm \Km \cDsq   \delpsisq \cxsq
 \bigg)^{\frac{1}{2}}.
\end{split}
\end{equation*}
Using this bound together with the results in \eqref{eq_ineq_totap_app_err} and \eqref{eq_cf_bnd_pfthm2}, the expected approximation error of observed signals is finally bounded as
\begin{equation*}
\begin{split}
&E\left[ \frac{1}{M} \sum_{m=1}^M \frac{1}{\Km} \sum_{i=1}^\Km \errtrainim(\psiest) \right] \\
&=
E\left[ \frac{1}{M} \sum_{m=1}^M \frac{1}{\Km} \sum_{i=1}^\Km
 \frac{1}{\Rim} \| \Sim \yim - \Sim \Dm(\psiest) \ximest \|
\right] \\
& = \frac{1}{M  \Theta( \Kc \Rc)}
\sum_{m=1}^M \sum_{i=1}^\Km
E\left[
\| \Sim \yim - \Sim \Dm(\psiest) \ximest \|
\right] \\
& \leq 
\frac{1}{M  \Theta( \Kc \Rc)}
\sqrt{\frac{\Ktot}{\eta_w} } \cf \\
& +
\frac{1}{M  \Theta( \Kc \Rc)}
 \sqrt{\Ktot} 
 \bigg(
  \sum_{m=1}^{M}
2  \sigma^2 \Km \Theta(\Rc) \\
&+
4 J \Nm \Km \cxsq 
+
4 J \Nm \Km \cDsq   \delpsisq \cxsq
 \bigg)^{\frac{1}{2}} \\
 & \leq
  \frac{\cf}{\sqrt{\eta_w M \Theta(\Kc)} \Theta(\Rc)} \\
  &+ \sqrt{\frac{2 \sigma^2}{\Theta(\Rc)}}
  +\frac{2 \cx (1+\cD \delpsi) \sqrt{J \Theta(N)}}{\Theta(\Rc)}
\end{split}
\end{equation*}
where we have used the relations $\Km = \Theta(\Kc)$, $\Rm= \Theta(\Rc)$, $\Nm = \Theta(\Nc)$, and $\Ktot = M\Theta(\Kc)$ to arrive at the statement of the theorem.
\end{proof}

\section*{Appendix D: Proof of Lemma \ref{lem_var_avgerrtrain}}

\begin{proof}

Recalling that $\{\ym\} \cup \{\yim\}_{i=1}^\Km  $ are i.i.d., as well as $\{\Sm\} \cup \{\Sim\}_{i=1}^\Km  $, we first observe that for a fixed, given $\psivect_0 \in \Psispace$, the coefficients $\{\xmest(\psivect_0) \} \cup \{ \ximest(\psivect_0) \}_{i=1}^\Km$ are i.i.d., from which it follows that the errors $\{\errtrainm(\psivect_0) \} \cup \{ \errtrainim(\psivect_0) \}_{i=1}^\Km$ are i.i.d.~as well. The variance of $\avgerrtrain(\psivect_0) $ is then given by

\begin{equation}
\label{eq_var_zpsi0_v1}
\begin{split}
\var(\avgerrtrain(\psivect_0) ) &= \var  \left( \frac{1}{M} \sum_{m=1}^M \frac{1}{\Km} \sum_{i=1}^\Km \errtrainim(\psivect_0) \right) \\
&=  \frac{1}{M^2} \sum_{m=1}^M \frac{1}{(\Km)^2}   \sum_{i=1}^\Km \var( \errtrainim(\psivect_0)) \\
&=  \frac{1}{M^2} \sum_{m=1}^M \frac{1}{\Km}   \var( \errtrainm(\psivect_0)).
\end{split}
\end{equation}
We next derive an upper bound for $\var( \errtrainm(\psivect_0))$. We have
\begin{equation}
\label{eq_var_empsi0_v1}
\begin{split}
&\var( \errtrainm(\psivect_0)) \leq E[ ( \errtrainm(\psivect_0) )^2 ]  \\
&= E\left[  \frac{1}{(\Rm)^2} \, \| \Sm \ym - \Sm \Dm(\psivect_0) \xmest(\psivect_0) \|^2   \right] .
\end{split}
\end{equation}
The definition of  $\xmest(\psivect_0)$ in \eqref{eq_defn_xmest_psi0} implies that
\begin{equation*}
\begin{split}
 & \eta_{w} \norm{ \Sm \ym- \Sm \Dm(\psivect_0) \xmest(\psivect_0)}^2 +  \eta_{x}  \norm{\xmest(\psivect_0)}_1 \\
 &\leq  \eta_{w} \norm{ \Sm \ym- \Sm \Dm(\psivect_0) \xm }^2 +  \eta_{x}  \norm{\xm}_1
\end{split}
\end{equation*}
where $\xm$ is the coefficient vector in the representation $\ym = \Dm(\psim) \xm + \wm$ of the graph signal $\ym$. This implies
\begin{equation*}
\begin{split}
& \norm{ \Sm \ym- \Sm \Dm(\psivect_0) \xmest(\psivect_0)}^2  \\
 &\leq  \norm{ \Sm \ym- \Sm \Dm(\psivect_0) \xm }^2 + \frac{ \eta_{x} }{\eta_{w} }  \norm{\xm}_1 \\
 &\leq  \norm{ \ym- \Dm(\psivect_0) \xm }^2 + \frac{ \eta_{x} }{\eta_{w} }  \norm{\xm}_1.
\end{split}
\end{equation*}
Using this inequality in \eqref{eq_var_empsi0_v1}, we get
\begin{equation}
\label{eq_var_empsi0_v2}
\begin{split}
&\var( \errtrainm(\psivect_0)) \leq \\
 &E\left[  \frac{1}{(\Rm)^2} \,
  \left(
   \| \ym - \Dm(\psivect_0) \xm \|^2 + \frac{ \eta_{x} }{\eta_{w} }  \norm{\xm}_1
  \right)
   \right] \\
 &=E\left[  \frac{1}{(\Rm)^2} \right] \,
 E \left[  \| \ym - \Dm(\psivect_0) \xm \|^2 + \frac{ \eta_{x} }{\eta_{w} }  \norm{\xm}_1 \right]
\end{split}
\end{equation}
where the equality is due to the independence of $\Sm$ (and hence $\Rm$) from  $\ym$ and $\xm$. Since the coefficients $\xm$ are assumed to be sampled from a Laplace distribution with scale parameter $\delta$, it follows that
\begin{equation}
\label{eq_exp_xm1}
E[ \| \xm \|_1 ] = J \Nm \delta.
\end{equation}
We proceed by deriving an upper bound for the expectation of the term $ \| \ym - \Dm(\psivect_0) \xm \|^2 $ in \eqref{eq_var_empsi0_v2}. It is easy to verify the inequality $ \| a+b \|^2 \leq 2 \|a\|^2 + 2 \| b \|^2$ for any two vectors $a, b \in \RN$.  We then have
\begin{equation}
\label{eq_ym_dmpsi0_xm_v1}
\begin{split}
&\| \ym - \Dm(\psivect_0) \xm \|^2  \\
&= \| \ym - \Dm(\psim) \xm  + \Dm(\psim) \xm  - \Dm(\psivect_0) \xm   \|^2  \\
& \leq  2  \| \ym - \Dm(\psim) \xm \|^2  + 2 \| \Dm(\psim) \xm  - \Dm(\psivect_0) \xm   \|^2 \\
& \leq 2 \| \wm \|^2 + 2 \|  \Dm(\psim) - \Dm(\psivect_0) \|_F^2 \ \| \xm \|^2.
\end{split}
\end{equation}
In order to derive an upper bound for the difference between the dictionaries $\Dm(\psim)$ and $\Dm(\psivect_0)$, let us first introduce the notation  $\gjpsi (\cdot)$ in order to refer to a kernel $\gj(\cdot)$ by particularly emphasizing the kernel parameter vector $\psivect$ it is generated from. We have
\begin{equation}
\label{eq_Dm_diff}
\begin{split}
& \|  \Dm(\psim) - \Dm(\psivect_0) \|_F^2 = \sum_{j=1}^J \|  \Djm (\psim)  -  \Djm (\psivect_0) \|_F^2 \\
&  = \sum_{j=1}^J \| \Um \gjpsim(\Lammm) (\Um)^T -  \Um \gjpsiz(\Lammm) (\Um)^T  \|_F^2 \\
&  = \sum_{j=1}^J  \| \gjpsim(\Lammm) - \gjpsiz(\Lammm)  \|_F^2 \\
&  = \sum_{j=1}^J  \sum_{n=1}^\Nm (\gjpsim(\lamnm) -  \gjpsiz(\lamnm) )^2 
\leq J \Nm
\end{split} 
\end{equation}
where $\lamnm$ refers to the $n$-th eigenvalue of the graph Laplacian $\Lapm$, and the inequality follows from the fact that the kernels $\gj$ are Gaussian functions taking values between $0$ and $1$. Using this bound in \eqref{eq_ym_dmpsi0_xm_v1} and taking expectation, we get
\begin{equation}
\label{eq_ym_dmpsi0_xm_v2}
\begin{split}
&E[ \, \| \ym - \Dm(\psivect_0) \xm \|^2  \,] 
 \leq 2 E[   \| \wm \|^2   ] + 2 J \Nm \ E[ \| \xm \|^2 ] \\
 &= 2 \Nm \sigma^2 + 4  (J \Nm \delta)^2 
\end{split}
\end{equation}
where the identities $E[   \| \wm \|^2   ] = \Nm \sigma^2 $ and $E[ \| \xm \|^2 ]= 2 J \Nm \delta^2 $ follow from the statistics of the normal and the Laplace distributions, respectively.

Returning back to the expression in \eqref{eq_var_empsi0_v2} and using the results \eqref{eq_exp_xm1} and \eqref{eq_ym_dmpsi0_xm_v2}, we obtain
\begin{equation*}
\begin{split}
&\var( \errtrainm(\psivect_0)) \\
  & \leq  E\left[  (\Rm)^{-2} \right] \,
 \left( E \left[  \| \ym - \Dm(\psivect_0) \xm \|^2 \right] + \frac{ \eta_{x} }{\eta_{w} } E[ \norm{\xm}_1 ] \right) \\
 & \leq E\left[  (\Rm)^{-2} \right] 
 (   2 \Nm \sigma^2  + 4 (J \Nm \delta)^2 +   \eta_{x} \eta_{w}^{-1}  J \Nm \delta ).
\end{split}
\end{equation*}
Finally, using this upper bound in the expression for $\var(\avgerrtrain(\psivect_0) ) $ in \eqref{eq_var_zpsi0_v1}, we arrive at the result stated in the lemma

\begin{equation*}
\begin{split}
\var(\avgerrtrain(\psivect_0) ) 
&=  \frac{1}{M^2} \sum_{m=1}^M \frac{1}{\Km}   \var( \errtrainm(\psivect_0)) \\
& \leq \frac{1}{M^2} \sum_{m=1}^M \frac{1}{\Km}  E\left[  (\Rm)^{-2} \right] 
 (   2 \Nm \sigma^2  \\
& \quad \quad 
+ 4 (J \Nm \delta)^2  
+ \eta_{x} \eta_{w}^{-1}  J \Nm \delta ) \\
& = \frac{1}{M^2} \sum_{m=1}^M  \frac{\cparm}{\Km}.
\end{split}
\end{equation*}
\end{proof}

\section*{Appendix E: Proof of Lemma \ref{lem_Dm_Lipsch}}

\begin{proof}

Similarly to \eqref{eq_Dm_diff}, we have
\begin{equation}
\label{eq_Dmdiff_lemma1}
\begin{split}
& \|  \Dm(\psitwo) - \Dm(\psione) \|_F^2   = \sum_{j=1}^J  \| \gjpsitwo(\Lammm) - \gjpsione(\Lammm)  \|_F^2 \\
&  = \sum_{j=1}^J  \sum_{n=1}^\Nm | \gjpsitwo(\lamnm) -  \gjpsione(\lamnm) |^2.
\end{split} 
\end{equation}
For a fixed $j$ and eigenvalue $\lamnm$,  let us consider $ \gjpsi(\lamnm) $ as a function of the vector $ \psivect \in \Psispace$. Then from the mean value theorem for functions of several variables, we have 
\begin{equation*}
\begin{split}
| \gjpsitwo(\lamnm) -  \gjpsione(\lamnm) | 
&\leq 
 \sup_{\psivect \in \Psispace}  \left \| \frac{\partial  \gjpsi(\lamnm) }{ \partial \psivect} \right  \|
 \  \| \psitwo - \psione \| \\
& \leq \cD \, \| \psitwo - \psione \|
\end{split}
\end{equation*}
where we define
\begin{equation*}
\begin{split}
\cD \triangleq \max_{j=1, \dots, J} \ \sup_{\psivect \in \Psispace, \lambda \in [0,2]}  \left \| \frac{\partial  \gjpsi(\lambda) }{ \partial \psivect} \right  \|.
\end{split}
\end{equation*}
Since $\Psispace $ is assumed to be a compact subset of $\R^{2J}$ that excludes $\psivect$ with $s_j =0$, and the kernels are continuous functions of $\psivect$, the supremum $\cD $ in the above expression exists and it is finite. Using this in \eqref{eq_Dmdiff_lemma1}, we get
\begin{equation*}
\begin{split}
\|  \Dm(\psitwo) - \Dm(\psione) \|_F^2 &\leq  \sum_{j=1}^J  \sum_{n=1}^\Nm \cDsq  \| \psitwo - \psione \|^2 \\
 & = J \Nm \cDsq \| \psitwo - \psione \|^2.
\end{split}
\end{equation*}
\end{proof}

\section*{Appendix F: Proof of Lemma \ref{lem_finite_mom_coeff}}

\begin{proof}
We need to show that the norms of the coefficient vectors have finite second moments. Let us begin with $\ximest(\psiest) $. Due to the optimality of $\ximest(\psiest) $ in \eqref{eq_defn_ximest}, we have 

\begin{equation*}
\begin{split}
& \eta_{x}  \|  \ximest(\psiest)  \|_1 +\eta_{w} \norm{ \Sim \yim- \Sim \Dmpsiest  \ximest(\psiest) }^2 \\
 & \leq 
 \eta_{x}  \|  \xim  \|_1 +\eta_{w} \norm{ \Sim \yim- \Sim \Dmpsiest  \xim }^2
\end{split}
\end{equation*}
where $\xim$ is the true coefficient vector generating the signal $\yim$ with respect to the model \eqref{eq_new_sig_model},  and $\ximest (\psiest)$ is its estimate via \eqref{eq_defn_ximest}. It follows that
\begin{equation*}
\begin{split}
  \|  \ximest(\psiest)  \|_1 
 \leq 
  \|  \xim  \|_1 + \frac{\eta_{w}}{ \eta_{x}} \norm{ \yim- \Dmpsiest  \xim }^2.
\end{split}
\end{equation*}
Plugging in the signal model $\yim = \Dmpsim \xim + \wim$ \eqref{eq_new_sig_model} in the above equation and squaring both sides we get
\begin{equation*}
\begin{split}
 & \|  \ximest(\psiest)  \|_1^2 \\
& \leq 
  \|  \xim  \|_1^2 
  +  \frac{2 \eta_{w}}{ \eta_{x}}   \|  \xim  \|_1  \norm{ ( \Dmpsim  - \Dmpsiest ) \xim + \wim }^2 \\
  &+ \frac{\eta_{w}^2}{ \eta_{x}^2} \norm{ ( \Dmpsim  - \Dmpsiest ) \xim + \wim }^4 
\end{split}
\end{equation*}
which gives
\begin{equation}
\label{eq_ineq_sec_mom_ximest}
\begin{split}
& E \left [\|  \ximest(\psiest)  \|^2 \right ]  \leq E \left [\|  \ximest(\psiest)  \|_1^2 \right ]  
 \leq
E \left [  \|  \xim  \|_1^2 \right  ] \\
&  +  \frac{2 \eta_{w}}{ \eta_{x}}  E \left [ \|  \xim  \|_1  \norm{ ( \Dmpsim  - \Dmpsiest ) \xim + \wim }^2 \right  ] \\
  &+ \frac{\eta_{w}^2}{ \eta_{x}^2} E \left [ \norm{ ( \Dmpsim  - \Dmpsiest ) \xim + \wim }^4 \right  ].
\end{split}
\end{equation}
We continue by showing the finiteness of each of the three rightmost terms in \eqref{eq_ineq_sec_mom_ximest}. Since the true coefficients $\xim$ have a Laplace distribution, they have finite moments, due to which the term $E \left [  \|  \xim  \|_1^2 \right  ]$ is finite. Next, the expression inside the expectation in the second term can be upper bounded as
\begin{equation}
\label{eq_sec_term_lem2}
\begin{split}
& \|  \xim  \|_1  \norm{ ( \Dmpsim  - \Dmpsiest ) \xim + \wim }^2 \\
& \leq  \|  \xim  \|_1 
 \left( 2 \norm{( \Dmpsim  - \Dmpsiest ) \xim}^2 + 2 \norm{\wim}^2 
  \right) \\
& \leq 
2  \, \| \Dmpsim  - \Dmpsiest \|_F^2 \, \|  \xim  \|_1^3 
+  2   \|  \xim  \|_1  \norm{\wim}^2 \\
& \leq  
 2 J \Nm \cDsq \, \|  \psim - \psiest \|^2 \, \|  \xim  \|_1^3 
 +  2   \|  \xim  \|_1  \norm{\wim}^2 \\
 & \leq
 2 J \Nm \cDsq \, \cpsidiffsq \, \|  \xim  \|_1^3 
 +  2   \|  \xim  \|_1  \norm{\wim}^2
\end{split}
\end{equation}
where the third inequality is due to Lemma \ref{lem_Dm_Lipsch}, and the constant in the fourth inequality is defined as $\cpsidiff \triangleq \sup_{\psione, \psitwo \in \Psispace} \|  \psione - \psitwo \| $, which is finite since $\Psispace$ is compact. Taking expectation in \eqref{eq_sec_term_lem2} and recalling the independence of $\xim $ and $\wim$, we obtain
\begin{equation*}
\begin{split}
& E \left[ \|  \xim  \|_1  \norm{ ( \Dmpsim  - \Dmpsiest ) \xim + \wim }^2 \right] \\
& \leq
 2 J \Nm \cDsq \, \cpsidiffsq \, E \left[ \|  \xim  \|_1^3 \right]
 +  2 E \left[   \|  \xim  \|_1 \right] E \left[  \norm{\wim}^2 \right]
 < \infty
\end{split}
\end{equation*}
since the Laplace and the Gaussian distributions have finite moments. Finally, for the third term in \eqref{eq_ineq_sec_mom_ximest} similar steps yield
\begin{equation*}
\begin{split}
& \norm{ ( \Dmpsim  - \Dmpsiest ) \xim + \wim } \\
& \leq 
\sqrt{ J \Nm \cDsq  \cpsidiffsq } \, \| \xim \| + \| \wim \|
\end{split}
\end{equation*}
from which 
\begin{equation*}
\begin{split}
& E \left[ \norm{ ( \Dmpsim  - \Dmpsiest ) \xim + \wim }^4 \right] \\
&\leq
\sum_{k=0}^4 { 4 \choose k}   (J \Nm \cDsq  \cpsidiffsq)^{\frac{k}{2}} \,
E \left[ \| \xim \|^k \right]  \, E\left[ \| \wim \|^{4-k} \right]
< \infty.
\end{split}
\end{equation*}
Hence, due to \eqref{eq_ineq_sec_mom_ximest} we conclude that $E \left [\|  \ximest(\psiest)  \|^2 \right ]$ is finite and upper bounded. The finiteness and upper boundedness of the second expression in the lemma are implied by those of the first one, observing that
\begin{equation*}
\begin{split}
E[\| \xim - \ximest(\psiest) \|^2] 
\leq 2 E[\| \xim  \|^2]   
+ 
2 E[\| \ximest(\psiest) \|^2].
\end{split}
\end{equation*}
The existence of the constant $\cx$ stated in the lemma then follows, which depends on the size parameters $J, \Nm$, the distribution parameters $\delta, \sigma$, and the domain $\Psispace$.
\end{proof}

\section*{Appendix G: Analysis of Remark 1}

\begin{proof}
Our approach will be to examine the probability expression in \eqref{eq_cor_exp_test_error} and first determine at which rate the accuracy parameters $\ctesterr $ and $\varepsilon$ may be allowed to vary with $M$ and $\Kc$ such that the probability in \eqref{eq_cor_exp_test_error}  remains fixed above 0. Obtaining an upper bound on the expected estimation error of unobserved samples in this manner, we can then compare the errors obtained in the multi-graph and the single-graph regimes. 

In Corollary \ref{cor_exp_test_error}, let us fix $0<\tau\ll 1$. We may model the covering number $\covnumcprobeps$ to be limited to some rate $\covnumcprobeps = O(\varepsilon^ {-\covnumrate})$ as the parameter $\varepsilon$ decreases, where the actual value of $\covnumrate>0 $ will not be critical for our analysis. Then, choosing the parameters  $\varepsilon$ and $\ctesterr $ as 
\begin{equation*}
\begin{split}
\varepsilon = \Theta \left( \frac{1}{(M \Kc)^{\frac{1}{\covnumrate+2}}} \right),
\quad
\ctesterr = \Theta(1) + \Theta \left( \frac{1}{\sqrt{M \Kc}} \right) + \Theta( \delpsi )
\end{split}
\end{equation*}
%
%
allows the terms involving $\varepsilon$ and $\ctesterr $ to be fixed to sufficiently small values, so that the probability expression in \eqref{eq_cor_exp_test_error} is fixed to a  positive value. In this case, from \eqref{eq_cor_testerr_acc_bound}, the expected estimation error of unobserved samples is upper bounded as
\begin{equation}
\label{eq_pf_rem1_testerr}
\begin{split}
&E \left[ \frac{1}{M} \sum_{m=1}^M  \frac{1}{\Km} \sum_{i=1}^\Km \errtestim(\psiest) \right] 
 \leq  \quad 
\ctesterr + \csmooth + (2+\cprob) \varepsilon \\
& = c_1 + \frac{c_2}{(M \Kc)^{\frac{1}{\covnumrate+2}}}
+ \frac{c_3}{\sqrt{M \Kc}} + c_4 \delpsi
\end{split}
\end{equation}
for some suitable constants $c_1$, $c_2$, $c_3$, and $c_4$. We proceed by comparing the error upper bounds in \eqref{eq_pf_rem1_testerr} for the cases of joint learning on multiple graphs ($M>1$) and individual learning on single graphs ($M=1$). This comparison is in favor of the multi-graph setting when
\begin{equation}
\label{eq_multi_graph_better_cond}
\begin{split}
\frac{c_2}{(M \Kc)^{\frac{1}{\covnumrate+2}}}
+ \frac{c_3}{\sqrt{M \Kc}} + c_4 \delpsi 
< 
\frac{c_2}{\Kc ^{\frac{1}{\covnumrate+2}}}
+ \frac{c_3}{\sqrt{\Kc}} 
\end{split}
\end{equation}
recalling that the spectral discrepancy will disappear ($\delpsi =0$) in the individual learning setting. Since
\begin{equation*}
\begin{split}
\frac{c_2}{(M \Kc)^{\frac{1}{\covnumrate+2}}} < \frac{c_2}{ \Kc ^{\frac{1}{\covnumrate+2}}}
\end{split}
\end{equation*}
for any $M>1$ and any $\covnumrate >0$, a sufficient condition that guarantees \eqref{eq_multi_graph_better_cond} is
\begin{equation*}
\begin{split}
\frac{c_3}{\sqrt{M \Kc}} + c_4 \delpsi 
<
\frac{c_3}{\sqrt{ \Kc}} 
\end{split}
\end{equation*}
which is equivalent to
\begin{equation*}
\begin{split}
\sqrt{ \Kc} < \frac{c_3}{c_4} \left( 1 - \frac{1}{\sqrt{M}} \right) \frac{1}{\delpsi}
\end{split}
\end{equation*}
or
\begin{equation*}
\begin{split}
\Kc < O\left( \frac{1}{\delpsisq} \right).
\end{split}
\end{equation*}
\end{proof}

\bibliographystyle{IEEEtran}
\bibliography{export}


%
%
%
%

\vfill

\end{document}